\documentclass[10pt,twocolumn,letterpaper]{article}

\usepackage{iccv}
\usepackage{times}
\usepackage{epsfig}
\usepackage{graphicx}
\usepackage{amsmath}
\usepackage{amssymb}

\usepackage{booktabs}
\usepackage{pifont}
\usepackage{color}
\usepackage{multirow}
\usepackage{overpic}
\usepackage{arydshln}
\usepackage{makecell}
\usepackage{times}
\usepackage{epsfig}
\usepackage[T1]{fontenc}
\usepackage{hhline}
\usepackage{xcolor}
\usepackage{pifont}
\usepackage{enumitem}
\usepackage{colortbl}
\usepackage{verbatim}
\usepackage{bm}
\usepackage{stfloats} 
\usepackage{lipsum}
\usepackage[pagebackref=true,breaklinks=true,letterpaper=true,colorlinks,bookmarks=false]{hyperref}

\usepackage[capitalize]{cleveref}
\crefname{section}{Sec.}{Secs.}
\Crefname{section}{Section}{Sections}
\Crefname{table}{Table}{Tables}
\crefname{table}{Tab.}{Tabs.}

\definecolor{darkpastelgreen}{rgb}{0.01, 0.75, 0.24}
\definecolor{darkpink}{rgb}{0.91, 0.33, 0.5}
\definecolor{mygray}{gray}{.92}
\definecolor{mygrayunder}{gray}{.98}
\definecolor{mygray1}{gray}{.50}
\definecolor{mygreen}{RGB}{93,174,86}
\definecolor{myred}{RGB}{245,55,45}
\definecolor{myorange}{RGB}{0,0,0}
\newcommand{\myPara}[1]{\textbf{#1}\quad}

\definecolor{improve}{RGB}{70,200,90}
\definecolor{linkcolor}{RGB}{255,0,0}
\definecolor{urlcolor}{RGB}{255,105,180}
\definecolor{citecolor}{RGB}{0, 80, 200}
\definecolor{citecolor1}{RGB}{0,153,255}
\usepackage{hyperref}
\newcommand\blfootnote[1]{%
  \begingroup
  \renewcommand\thefootnote{}\footnote{#1}%
  \addtocounter{footnote}{-1}%
  \endgroup
}
\hypersetup{colorlinks=true,linkcolor=linkcolor,urlcolor=urlcolor,citecolor=citecolor1}

\usepackage[breaklinks=true,bookmarks=false]{hyperref}

\iccvfinalcopy 

\def\iccvPaperID{****} 

\ificcvfinal\fi

\begin{document}

\title{Spatial-Aware Token for Weakly Supervised Object Localization}

\author{Pingyu Wu$^{1}$, Wei Zhai$^{1,\dagger}$, Yang Cao$^{1,3}$, Jiebo Luo$^{2}$, Zheng-Jun Zha$^{1}$\\
{$^{1}$~University of Science and Technology of China} \qquad
{$^{2}$~University of Rochester}\\
{$^{3}$~Institute of Artificial Intelligence, Hefei Comprehensive National Science Center}\\
\small{\texttt{\{wpy364755620@@mail., wzhai056@mail., forrest@\}ustc.edu.cn}} \\
\small{\texttt{jluo@cs.rochester.edu}} \qquad
\small{\texttt{zhazj@ustc.edu.cn}}
 }

\maketitle
\ificcvfinal\thispagestyle{empty}\fi
\blfootnote{$\dagger$ Corresponding author.}


\begin{abstract}
Weakly supervised object localization (WSOL) is a challenging task aiming to localize objects with only image-level supervision. Recent works apply visual transformer to WSOL and achieve significant success by exploiting the long-range feature dependency in self-attention mechanism. However, existing transformer-based methods synthesize the classification feature maps as the localization map, which leads to optimization conflicts between classification and localization tasks. To address this problem, we propose to learn a task-specific spatial-aware token (SAT) to condition localization in a weakly supervised manner. Specifically, a spatial token is first introduced in the input space to aggregate representations for localization task. Then a spatial aware attention module is constructed, which allows spatial token to generate foreground probabilities of different patches by querying and to extract localization knowledge from the classification task. Besides, for the problem of sparse and unbalanced pixel-level supervision obtained from the image-level label, two spatial constraints, including batch area loss and normalization loss, are designed to compensate and enhance this supervision. Experiments show that the proposed SAT achieves state-of-the-art performance on both CUB-200 and ImageNet, with 98.45\% and 73.13\% GT-known Loc, respectively. Even under the extreme setting of using only 1 image per class from ImageNet for training, SAT already exceeds the SOTA method by 2.1\% GT-known Loc. Code and models are available at \href{https://github.com/wpy1999/SAT}{\color{magenta}https://github.com/wpy1999/SAT}.

\end{abstract}

\begin{figure}[t]
\centering
\begin{overpic}[width=1\linewidth]{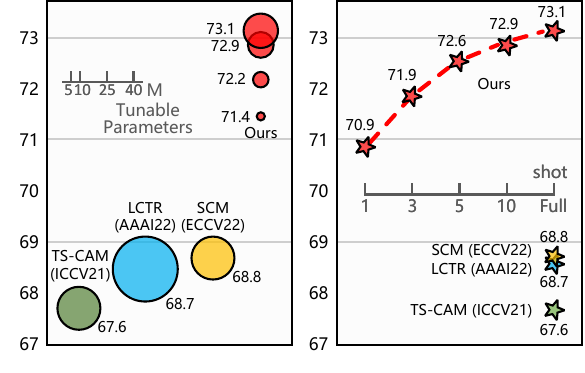}
\put(11., 1.){{\textbf{Tunable Parameters}}}
\put(62.2, 1.){{\textbf{Few-shot learning}}}
\put(-0.1, 20){\rotatebox{90}{\small\textbf{GT-known Loc}}}
\end{overpic}
    \caption{\textbf{GT-known Loc on ImageNet}. The proposed method only needs to fine-tune a small number of parameters or requires a few training images to achieve significant improvement.}
    \label{fig:fig1}
\end{figure}

\section{Introduction}
Weakly supervised object localization (WSOL) aims to localize objects with only image-level labels available. Since no expensive bounding box or pixel-level annotations are required, WSOL significantly reduces the cost of manual annotations~\cite{shaharabany2022looking,zhang2021weakly,kim2021normalization,lee2022anti,choe2020attention,zhang2020rethinking,singh2017hide,zhang2018adversarial} and has attracted increasing attention in the research community~\cite{yun2019cutmix,zhai2022one,zhai2022exploring,yang2022deep,jing2021amalgamating,wang2020dual,wang2021structured,wang2022semantic,liu2019adaptive,yang2019making,tan2020learning}.

As a representative work, CAM~\cite{zhou2016learning} extracts class activation maps from the classifier as localization maps. However, CAM is usually coarse and focuses on the most discriminative regions, leading to imprecise and incomplete localization results. To solve these problems, many CNN-based methods have been proposed, such as adversarial erasure~\cite{singh2017hide,zhang2018adversarial,choe2019attention,mai2020erasing}, divergent activation~\cite{xue2019danet,yun2019cutmix,singh2017hide}, seed region growing~\cite{wei2021shallow,zhang2018self}, regularization~\cite{pan2021unveiling,lee2022anti,zhang2020inter}, feature refining~\cite{bae2020rethinking,yang2020combinational,wei2021shallow}, regression-based~\cite{xu2022cream,zhang2020rethinking,guo2021strengthen}.

Recently, transformer~\cite{dosovitskiy2020image,touvron2021training} has been introduced to the field of computer vision with great success. Benefiting from the long-range feature dependency, the attention map upon the class token usually can capture the object region relevant to classification in the whole image, thus it is widely used for localization in the transformer-based WSOL methods. The pioneering work TS-CAM~\cite{gao2021ts} accumulates attention maps of each layer and mixes them with semantic-aware maps to achieve localization task. Further, to address the problem that transformer lacks inherent spatial coherence of the object, LCTR~\cite{chen2021lctr} and SCM~\cite{bai2022weakly} propose to consider cross-patch information and use activation diffusion to increase the local continuity of localization map, respectively.

Existing transformer-based methods synthesize the feature maps learned by the classification task, such as attention maps, as the localization map, and try to increase its connectivity and completeness. However, this convenient approach results in optimization conflicts between classification and localization tasks. 1) For classification, making classification feature maps learn more object regions with low discrimination will reduce the classification ability. 2) For localization, the learning of the localization map is limited by the properties of the feature maps. For example, the attention map is generated by the softmax function, which is difficult to produce a balanced and comprehensive response over the object. Therefore, to achieve a more promising localization performance and avoid optimization conflicts, it is necessary to construct task-specific parameters for the generation and learning of localization map.

Based on transformer architecture, a straightforward idea is to introduce a spatial token in the input space to aggregate global representation for the localization task. And the localization map can be further obtained by interacting the spatial token with each patch in the forward propagation. To this end, we propose a Spatial-Query Attention (SQA) module that takes the spatial token as a query to calculate the similarity with different patches and produces the localization map efficiently. Meanwhile, to make the spatial-aware token obtain localization supervision from image-level labels, the localization map is treated as a visual cue to participate in the calculation of cross attention, thus the generation and learning of localization map can be well adapted to the transformer-based classification model.

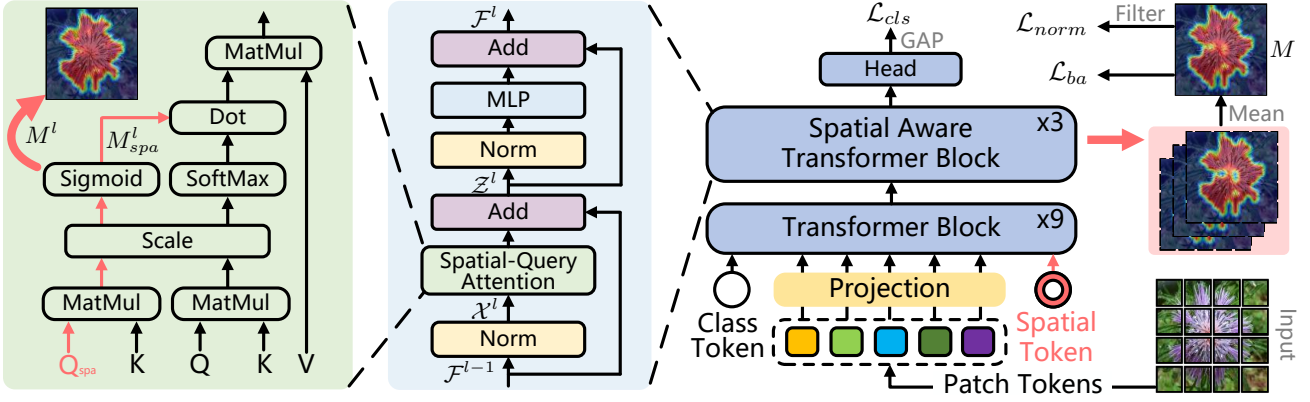
\begin{figure*}[t]
\centering
    \begin{overpic}[width=0.99\linewidth]{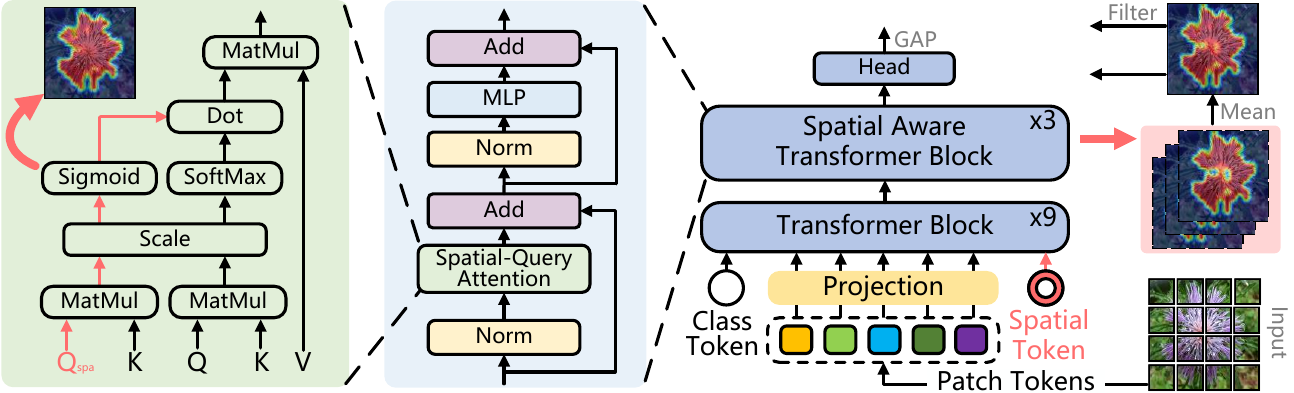}
    \put(34, 1.){\small\textbf{$\mathcal{F}^{l-1}$}}
    \put(36.1, 5.8){\small\textbf{$\mathcal{X}^{l}$}}
    \put(36.1, 15.55){\small\textbf{$\mathcal{Z}^{l}$}}
    \put(36.1, 28.3){\small\textbf{$\mathcal{F}^{l}$}}
    \put(8, 19){\small\textbf{$M^{l}_{spa}$}}
    \put(1.8, 19.4){\small\textbf{$M^{l}$}}
    \put(97.5, 26){\small\textbf{$M$}} 
    \put(66.5, 28.8){\textbf{$\mathcal{L}_{cls}$}} 
    \put(80.5, 24.2){\textbf{$\mathcal{L}_{ba}$}} 
    \put(78, 28){\textbf{$\mathcal{L}_{norm}$}} 

    \end{overpic}
   \caption{\textbf{Framework} of SAT. It includes three spatial aware transformer blocks at the end of the network. Each block generates a localization map $M^{l}$ using the spatial-query attention module. The final localization map $M$ is obtained by fusing $M^{l}$ of different layers.}
    \label{fig:network}
\end{figure*}

Building upon this efficient SQA module, we propose a simple but effective \textbf{S}patial-\textbf{A}ware \textbf{T}oken (SAT) approach, as shown in Fig.~\ref{fig:network}. Specifically, a task-specific spatial token is introduced in the input space. To achieve the localization task, several SQA modules are applied in the transformer to produce localization maps learned from different layers and aggregate them together. In this way, the final localization map can maximally capture the localization knowledge from the whole classification model. However, the pixel-level supervision generated by image-level labels is sparse and unbalanced. To compensate and strengthen this supervision, we propose two spatial constraints, including batch area loss and normalization loss. Batch area loss aims to provide a sparse area supervision with prior knowledge to compensate for the insufficient supervision. Normalization loss is employed to enhance pixel-level supervision by encouraging the localization map to be more discriminatory.

Benefiting from the task-specific spatial token and avoiding optimization conflicts, SAT releases the potential of the model and achieves excellent performance in both classification and localization tasks. Besides, unlike synthetic classification feature maps as localization results, the generation and learning of localization map in SAT rely mainly on an extra lightweight token, which brings advantages of data-efficiency and tuning-efficiency. As illustrated in Fig.~\ref{fig:fig1}, SAT outperforms SOTA method SCM by \textbf{2.1\%} GT-known Loc with less than \textbf{0.1\%} training data on ImageNet. 

In summary, the contributions of this paper include:

\textbf{1)} We propose to learn a task-specific spatial token to implement the localization task, instead of synthesizing classification feature maps, thus avoiding optimization conflicts between classification and localization tasks.

\textbf{2)} We propose a simple but effective spatial-aware token (SAT) pipeline for WSOL, which utilizes a spatial-query attention (SQA) module to generate localization map through a spatial token and supervise it with two spatial constraints, including batch area loss and normalization loss.
	
\textbf{3)} Extensive experiments show that SAT outperforms SOTA methods by a large margin on multiple benchmarks and performs excellently even under extreme settings.

\section{Related work}
\myPara{CNN-based methods.} 
WSOL aims to learn the localization of objects with only image-level labels. CAM~\cite{zhou2016learning} proposes to synthesize depth feature maps with fully connected weights to obtain class activation maps (CAM). To increase the localization ability of CAM, HaS~\cite{singh2017hide} adopts a random erasure strategy forcing the network to attend to different object regions. ACoL~\cite{zhang2018adversarial} and EIL~\cite{mai2020erasing} use adversarial erasing to achieve learning of complementary regions. ADL~\cite{choe2020attention} drops the most discriminative or highlighted regions in the different layers during forward propagation. Besides, SPA~\cite{pan2021unveiling} designs a post-process approach to extract structure-preserving localization map from the feature maps. SPG~\cite{zhang2018self} and SPOL~\cite{wei2021shallow} facilitate learning of localization by selecting and spreading confidence regions as seeds. PSOL~\cite{zhang2020rethinking} and 
C$^{2}$AM~\cite{{xie2022contrastive}} divide WSOL into independent classification task and class-agnostic object localization. I2C~\cite{zhang2020inter} and ISIC~\cite{wei2022weakly} consider feature similarities across different objects to achieve more complete and robust localization. ORNet~\cite{xie2021online}, FAM~\cite{meng2021foreground} and BAS~\cite{wu2021background} propose to generate a foreground prediction map (FPM) to implement localization. Unlike FPM-based methods that require complex and heavy structures, and the learning of generator relies on a specific feature layer of the classification network. Our method simply and efficiently obtains localization maps from different layers by a task-specific token.

\myPara{Transformer-based methods.} 
Different from CNNs, transformer~\cite{dosovitskiy2020image,vaswani2017attention} can capture global cues with the advantage of long-range dependency in the self-attention mechanism, effectively alleviating the partial activation problem. TS-CAM~\cite{gao2021ts} mixes the attention maps of different layers and combines them with semantic-aware tokens to produce semantic-aware localization results. Based on TS-CAM, to address the lack of spatial consistency in transformer, LCTR~\cite{chen2021lctr} proposes to enhance the local perception capability among long-range feature dependencies by considering cross-patch information. SCM~\cite{bai2022weakly} increases the spatial coherence of attention maps by diffusing the semantic and spatial connections between patch tokens. Notably, they all generate localization map by synthesizing feature maps. Instead, we produce it by a task-specific spatial token.

\section{Methodology}

\subsection{Overview}
The overall network architecture of SAT is shown in Fig. ~\ref{fig:network}. Given an input image $I\in \mathbb{R}^{h \times w \times 3}$, a plain vision transformer splits it into a sequence of non-overlapping patches. Then the divided
patches are flattened and transformed into patch tokens $\{x_{n}\in \mathbb{R}^{1 \times D}, n=1, 2,...,H\times W \}$ by linear projection, where $H=h/P$, $W=w/P$, $P$ is the patch size, and $D$ is the number of channels. For simplicity, we omit the description of the batch size $B$. After grouping the class token and the extra spatial-aware token $x_{spa}$ with patch tokens, this token sequence $\mathcal{F}^{0}\in \mathbb{R}^{(N+2) \times D}$ is sent into stacked transformer blocks and spatial aware transformer blocks for subsequent representation learning.

The proposed spatial aware transformer block inherits the typical transformer block structure. Thus the entire network is characterized by alternating attention layers and MLP layers. For a given input token subsequent $\mathcal{F}^{l-1}$, the overall equations of the $l$-th block are defined as follows:
\begin{align}
&\mathcal{X}^{l}=\mathrm{Layer}\text{-}\mathrm{Norm}\Big(\mathcal{F}^{l-1}\Big), \\
&\mathcal{Z}^{l}=\mathcal{F}^{l-1} + \mathrm{Attention}\Big(\mathcal{X}^{l}\Big), \\
&\mathcal{F}^{l}=\mathcal{Z}^{l} +\mathrm{MLP}\Big(\mathrm{Layer}\text{-}\mathrm{Norm}\Big(\mathcal{Z}^{l}\Big)\Big).
\end{align}
\subsection{Spatial Aware Transformer Block}
The proposed spatial aware transformer block mainly contains a spatial-query attention (SQA) module to generate localization map and extract localization knowledge. Besides, two spatial constraints, including batch area loss $\mathcal{L}_{ba}$, and normalization loss $\mathcal{L}_{norm}$ are designed as the complement to the insufficient localization supervision.


\begin{figure*}[t]
\centering
\small
	\begin{overpic}[width=1\linewidth]{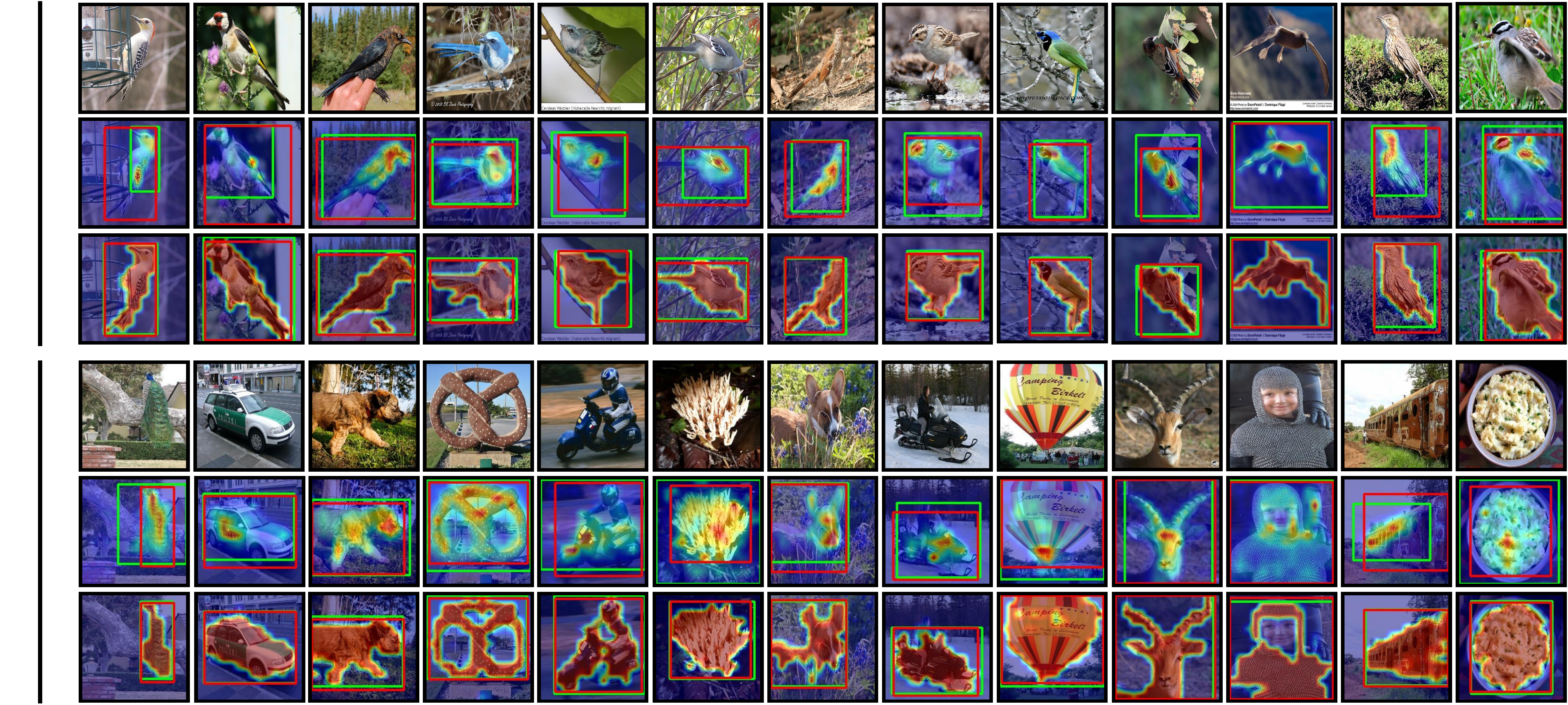}
	    \put(0.6, 7.3){\rotatebox{90}{\small\textbf{ImageNet}}}
	    \put(0.6, 30.2){\rotatebox{90}{\small\textbf{CUB-200}}}
	    \put(3.1, 2.1){\rotatebox{90}{\small\textbf{Ours}}}
	    \put(3.1, 7.4){\rotatebox{90}{\small\textbf{TS-CAM}}}
	    \put(3.1, 16.4){\rotatebox{90}{\small\textbf{Image}}}
	    
	    \put(3.1, 24.7){\rotatebox{90}{\small\textbf{Ours}}}
	    \put(3.1, 30.3){\rotatebox{90}{\small\textbf{TS-CAM}}}
	    \put(3.1, 39.1){\rotatebox{90}{\small\textbf{Image}}}
\end{overpic}
\caption{\textbf{Visualization comparison.} The ground-truth bounding boxes are in \textcolor{red}{red}, and the predicted bounding boxes are in \textcolor{green}{green}.}
\label{experiment_visual}
\end{figure*}
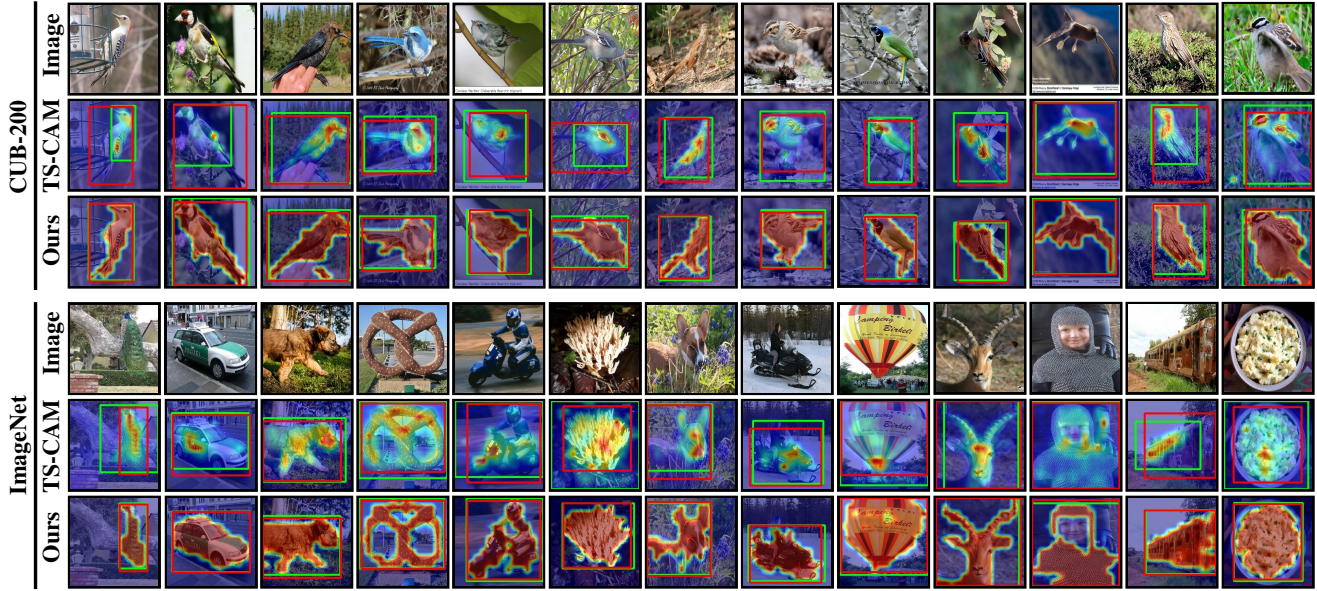

\myPara{Spatial-query Attention.}
The proposed SQA module is based on self-attention module and aims at converting the spatial-aware token $x_{spa}$ into a localization map $M^{l}_{spa}$ by querying different patches. Meanwhile, the generated $M^{l}_{spa}$ is applied to the cross-attention calculation in the form of dot product, thus obtaining localization supervision from the image-level labels. Specifically, the SQA module first linearly projects the token sequence $\mathcal{X}^{l-1}$ to the query matrix $\mathcal{Q}$, key matrix $\mathcal{K}$, value matrix $\mathcal{V}$, where $\mathcal{Q}$, $\mathcal{K}$, $\mathcal{V}$ $\in \mathbb{R}^{(N+2) \times D}$. Then the query vector corresponding to the spatial token $\mathcal{Q}_{spa}\in \mathbb{R}^{1 \times D}$ is selected from $\mathcal{Q}$ and used to query $\mathcal{K}$ to obtain the query results for each token. The similarity map between the spatial query vector and key matrix in the $l$-th layer is calculated as follows:
\begin{equation}
\mathcal{S}^{l}(\mathcal{Q}_{spa},\mathcal{K}) = \frac{\mathcal{Q}_{spa}\mathcal{K}^{\text{T}}}{\sqrt{D}}\in \mathbb{R}^{1 \times (N+2)},
\end{equation}
where $\sqrt{D}$ is employed as a scaling factor. After applying sigmoid activation, the similarity map is transformed into foreground probability $M^{l}_{pro}$ with a 0 to 1 interval distribution. This process can be expressed as follows:
\begin{equation}
M^{l}_{spa} = \mathrm{Sigmoid}(\mathcal{S}^{l}(\mathcal{Q}_{spa},\mathcal{K}))\in \mathbb{R}^{1 \times (N+2)}.
\end{equation}
The generated $M^{l}_{spa}$ is served as a visual cue to point out the object region and combined in the calculation of attention with the form of dot product. In this way, the whole SQA module can be formulated in the following form:
\begin{equation}
\mathrm{SQA}(\mathcal{X}^{l}) = \mathrm{Softmax}(\frac{\mathcal{Q}\mathcal{K}^{\text{T}}}{\sqrt{D}})*M^{l}_{spa}\mathcal{V},
\end{equation}
where $*$ denotes element-wise multiplication.

By cropping and reshaping, the part of the $M^{l}_{spa}$ to the patch tokens can be transformed into a localization map $M^{l}\in \mathbb{R}^{H \times W}$. After obtaining $M^{l}$ learned from different layers, we take their averages as the final localization map $M$ to increase the completeness and robustness. Although this approach can maximally capture the localization information in the classification network, the pixel-level supervision obtained from image-level labels is sparse and unbalanced. To compensate and strengthen this supervision, we design batch area loss and normalization loss.

\myPara{Batch Area.}
To alleviate the insufficient localization supervision caused by the absence of dense labels in WSOL, we propose to use batch area loss as a complement to localization supervision. It aims to constrain the average area of localization maps $\{M_{b}\in \mathbb{R}^{H \times W}, b=1, 2,...,B \}$ within a batch $B$ to a hyperparameter $\lambda$, as follows: 
\begin{equation}
    \mathcal{L}_{ba} =  \lvert\sum_{b}^B\sum_{i}^{H}\sum_{j}^{W}{( \lambda - \frac{M_{b}(i,j)}{B \times H \times W} )}\rvert,
    \label{eq:batch_area}
\end{equation}
where $\lambda$ is a sparse area supervision with prior knowledge. The $\lambda$ is set to 0.25 and 0.35 on CUB-200 and ImageNet.

\begin{table*}[t]
\renewcommand{\arraystretch}{1.}
\renewcommand{\tabcolsep}{7.5pt}
\begin{center}
\small
\begin{tabular}{l|c|c|c|c|c|c|c|c}
\Xhline{2.0\arrayrulewidth}
\hline
\multirow{2}{*}{\textbf{Methods}} & \multirow{2}{*}{\textbf{Venue}} & \multirow{2}{*}{\textbf{Backbone}} &  \multicolumn{3}{c|}{\textbf{CUB-200}~\textbf{\cite{wah2011caltech} Loc Acc.}} & \multicolumn{3}{c}{\textbf{ImageNet}~\textbf{\cite{russakovsky2015imagenet} Loc Acc.}} \\
\cline{4-9}
      & &  & \textbf{Top-1} & \textbf{Top-5} & \textbf{GT-known}  & \textbf{Top-1} & \textbf{Top-5} & \textbf{GT-known} \\
\hline
\Xhline{2.0\arrayrulewidth}
CAM~\cite{zhou2016learning} & CVPR$16$ &VGG16
& 41.06 & 50.66 & 55.10 & 42.80 & 54.86 & 59.00   \\
ORNet~\cite{xie2021online} & ICCV$21$ &VGG16
& 67.73 & 80.77 & 86.20 & 52.05 & 63.94 & 68.27 \\
BAS~\cite{wu2021background} & CVPR$22$ &VGG16
& 71.33 & 85.33 & 91.07 & 52.96 & 65.41  & 69.64 \\
Kim et al.~\cite{kim2022bridging} & CVPR$22$  &VGG16
& 70.83 & 88.07 & 93.17 & 49.94 & 63.25 & 68.92\\
CREAM~\cite{xu2022cream}  & CVPR$22$ & VGG16
& 70.44 &85.67 & 90.98 & 52.37 & 64.20 & 68.32 \\
\hline
GCNet~\cite{lu2020geometry} & ECCV$20$ &InceptionV3
& 58.58 & 71.00 & 75.30 & 49.06 & 58.09 & $-$\\
SPA~\cite{pan2021unveiling} & CVPR$21$ &InceptionV3
& 53.59 & 66.50 & 72.14 & 52.73 & 64.27 & 68.33 \\
FAM~\cite{meng2021foreground} & ICCV$21$ &InceptionV3
& 70.67 & $-$ & 87.25 & 55.24 & $-$ & 68.62 \\
CREAM~\cite{xu2022cream} & CVPR$22$ &InceptionV3 & 71.76 & 86.37 & 90.43 & 56.07 & 66.19 & 69.03 \\
BAS~\cite{wu2021background} & CVPR$22$ &InceptionV3
& 73.29 & 86.31 & 92.24 & 58.51 &\underline{69.00} & 71.93 \\
BagCAMs~\cite{zhu2022bagging} & ECCV$22$ & InceptionV3  & 60.07  & $-$ & 89.78 & 53.87  & $-$  & 71.02 \\
\hline
SPOL~\cite{wei2021shallow} & CVPR$21$&ResNet50
& 80.12 & 93.44 & 96.46 & 59.14 & 67.15 & 69.02 \\
DA-WSOL~\cite{zhu2022weakly} & CVPR$22$ &ResNet50
& 66.65 & $-$ & 81.83 & 55.84 & $-$ & 70.27 \\
BAS~\cite{wu2021background} & CVPR$22$ &ResNet50
& 77.25 & 90.08 & 95.13 & 57.18 & 68.44 & 71.77 \\
Kim et al.~\cite{kim2022bridging} & CVPR$22$  &ResNet50
& 73.16 & 86.68 & 91.60 & 53.76 & 65.75 & 69.89\\
CREAM~\cite{xu2022cream} & CVPR$22$ & ResNet50
& 76.03 & $-$ & 89.88 & 55.66 & $-$ &  69.31 \\
BagCAMs~\cite{zhu2022bagging} & ECCV$22$ & ResNet50  & 69.67  & $-$ & 94.01 & 44.24  & $-$  & \underline{72.08} \\
ISIC~\cite{wei2022weakly} & ECCV$22$ & ResNet50  & \underline{80.68} & \underline{94.08} & \underline{97.32} & \underline{59.61}  & 67.84  & 70.01 \\
\hline
TS-CAM~\cite{gao2021ts} & ICCV$21$ &Deit-S
& 71.30 & 83.80 & 87.70 & 53.40 & 64.30 & 67.60 \\
LCTR~\cite{chen2021lctr} & AAAI$22$ &Deit-S
& 79.20 & 89.90 & 92.40 & 56.10 & 65.80 & 68.70 \\
SCM~\cite{bai2022weakly} & ECCV$22$ &Deit-S
& 76.40 & 91.60 & 96.60 & 56.10 & 66.40 & 68.80 \\
\hline
\rowcolor{mygray}
\rowcolor{mygray}
SAT ({\color{red} ours}) & This Work & Deit-S & \textbf{80.96} & \textbf{94.13} & \textbf{98.45} & \textbf{60.15} & \textbf{70.52} & \textbf{73.13}\\ 

\hline
\Xhline{2.0\arrayrulewidth}
\end{tabular}
\end{center}
\vspace{-2mm}
\caption{\textbf{Comparison with state-of-the-art methods.} The best results are highlighted in \textbf{bold}, second are \underline{underlined}.}
\label{table:comparsion}
\end{table*}

\myPara{Normalization.} For unbalanced pixel-level supervision, a normalization loss is applied to enhance the pixel-level supervision based on the pixel response intensity. It encourages pixels to be more distinguishable about whether they are foreground or background, as presented by Eq.~\ref{eq:norm}. Before penalizing the uncertainty value of the localization map $M$, to further enhance its local consistency, we adopt a gaussian filter with a kernel size of 3x3 to establish spatial connections between adjacent patches. The standard deviation of gaussian kernel is set to 6 on all datasets. After applying gaussian filtering, the value of each patch is redistributed according to the adjacent patch values, and M is converted to $M^*$. By computing the normalization loss with $M^*$ can further encourage $M$ to have local consistency and avoid noise. It can be expressed as follows:
\begin{equation}
    \mathcal{L}_{norm} = \frac{1}{H \times W}\sum_{i}^{H}\sum_{j}^{W}{ M^*(i,j)(1-M^*(i,j))}.
    \label{eq:norm}
\end{equation}
\subsection{Total loss}
We follow the design of classification head in the baseline method TS-CAM~\cite{gao2021ts} to obtain the predicted probability $y$ and the cross-entropy classification loss ${L}_{cls}$. By jointly optimizing classification loss, batch area loss, and normalization loss, the total loss function is defined as follows:
\begin{equation}
    \label{eq:loss}
    \mathcal{L}=\mathcal{L}_{cls} + \mathcal{L}_{ba} + \mathcal{L}_{norm},
\end{equation}
where no extra hyperparameter is used to balance each loss.

\section{Experiment}

\subsection{Experimental Setup}
\myPara{Datasets.}
We evaluate the proposed SAT on three popular datasets, including \textbf{CUB-200}~\cite{wah2011caltech}, \textbf{ImageNet}~\cite{russakovsky2015imagenet}, \textbf{OpenImages}~\cite{choe2020evaluating}. CUB-200 is a fine-grained dataset with 200 bird species, which consists of 5,994 images for training and 5,794 images for testing. ImageNet contains about 1.2 million training images and 50,000 validation images from 1,000 different classes. OpenImages contains 100 categories, it has 29,819, 2,500, and 5,000 samples in training, val, and test sets, respectively. Apart from the image-level labels, both CUB-200 and OpenImages provide pixel-level mask labels, which are only used in the testing phase.

\myPara{Metrics.}
Following~\cite{wu2021background,pan2021unveiling,zhu2022weakly}, for localization, we utilize GT-known localization accuracy (\textbf{GT-known Loc}), Top-1/Top5 localization accuracy (\textbf{Top-1/Top-5 Loc}), and maximal box accuracy (\textbf{MaxBoxAccV2})~\cite{choe2020evaluating} as evaluation metrics. GT-known Loc is correct indicating that the intersection over union (\textbf{IoU}) of the predicted bounding box and the ground-truth bounding box is 50\% or more. Top-1/Top-5 Loc is correct when the ground-truth class belongs to Top-1/Top-5 prediction categories and GT-known Loc is correct. For mask, the peak intersection over union (\textbf{pIoU})~\cite{zhang2020rethinking} and pixel average precision (\textbf{PxAP})~\cite{choe2020evaluating} are adopted as metrics.

\begin{table}[t] 
\begin{center}
    \begin{minipage}{0.68\linewidth}
    
    \renewcommand{\arraystretch}{1.}
    \renewcommand{\tabcolsep}{1pt}
    \centering
    \small
    \begin{tabular}{c|cccc|ccc}
    \Xhline{2.\arrayrulewidth}
    \hline 
     & \multicolumn{4}{c|}{\textbf{Setting}} & \multicolumn{3}{c}{\textbf{Loc Acc.}} \\
    \hline
     & $\mathcal{L}_{cls}$ & $\mathcal{L}_{ba}$  & $\mathcal{L}_{norm}$ & $F.$ & \textbf{Top-1} & \textbf{Top-5} & \textbf{GT-k.}\\
    \Xhline{2.\arrayrulewidth}
    \hline 
    \textbf{(a)} & $\checkmark$ & & & & 56.86 & 66.67 & 69.29 \\
    \textbf{(b)} & $\checkmark$ & $\checkmark$ & & & 58.21 & 68.36 & 71.07 \\
    \textbf{(c)} & $\checkmark$ & $\checkmark$ & $\checkmark$ &  & 59.86 & 70.22 & 72.78\\
    \hline
    \rowcolor{mygray}
    \textbf{(d)} & $\checkmark$ & $\checkmark$ &  $\checkmark$ &  $\checkmark$ &\textbf{60.15} & \textbf{70.52} & \textbf{73.13}\\
    \hline
    \Xhline{2.\arrayrulewidth}
    \end{tabular}
    
    \end{minipage}
    \begin{minipage}{0.31\linewidth}
    \begin{overpic}[width=1.\linewidth]{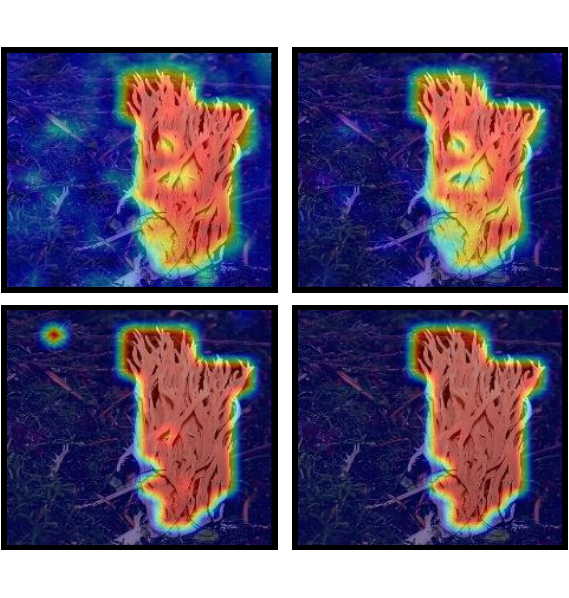}
        \put(19, 1){\textbf{\scriptsize{(c)}}}
	    \put(66, 1){\textbf{\scriptsize{(d)}}}
	    \put(19, 94){\textbf{\scriptsize{(a)}}}
	    \put(66, 94){\textbf{\scriptsize{(b)}}}
    \end{overpic}
    
    \end{minipage}
    \end{center}
    \vspace{-2mm}
    \caption{\textbf{Ablation study} of SAT on ImageNet. $F.$: Gaussian filter.}
    \label{table:Ablation}
\end{table}

\begin{figure}[t]
\begin{center}
    \begin{overpic}[width=1\linewidth]{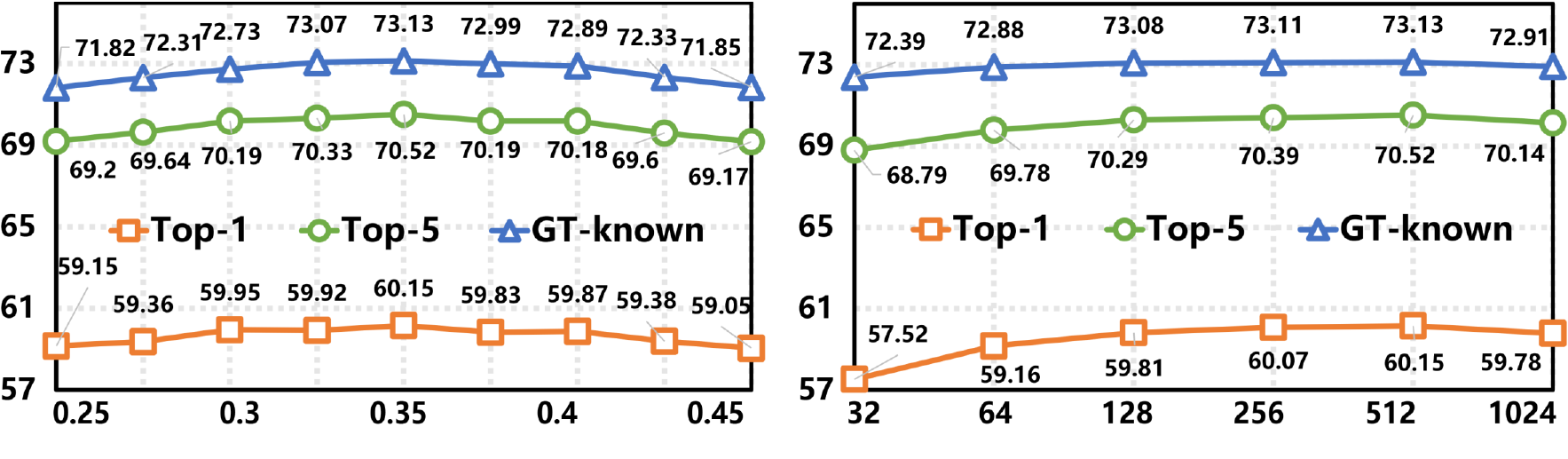}
	    
	    \put(22, -1){\textbf{\small{(a) $\lambda$}}}
	    \put(65,-1){\textbf{\small{(b) batch size}}}
	    
    \end{overpic}
    \end{center}
    \vspace{-2mm}
    \caption{\textbf{Hyperparameters.} Sensitivity analysis of hyperparameters (a) $\lambda$ and (b) batch size in $\mathcal{L}_{ba}$ on ImageNet.}
    \label{fig:ablation}
\end{figure}

\begin{table*}[t]
\begin{center}
\small 
\renewcommand{\arraystretch}{1}
\renewcommand{\tabcolsep}{4.5pt}
\begin{tabular}{l|c|c|c|cccc|cccc}
\Xhline{2.\arrayrulewidth}
\hline 
\multirow{2}{*}{\textbf{Method}} & \multirow{2}{*}{\textbf{Venue}} & \multirow{2}{*}{\textbf{Backbone}} & \multirow{2}{*}{\textbf{Resolution}} &  \multicolumn{4}{c|}{\textbf{CUB-200}~\textbf{\cite{wah2011caltech}}} & \multicolumn{4}{c}{\textbf{ImageNet}~\textbf{\cite{russakovsky2015imagenet}}} \\
 & & & & \textbf{$\delta$ = 0.3} & \textbf{$\delta$ = 0.5} & \textbf{$\delta$ = 0.7} & \textbf{Mean} & \textbf{$\delta$ = 0.3} & \textbf{$\delta$ = 0.5} & \textbf{$\delta$ = 0.7} & \textbf{Mean} \\
\hline 
\Xhline{2.\arrayrulewidth}
$\text{Kim et al.}$~\cite{kim2022bridging} & CVPR$22$ & ResNet50 & 28$\times$28 & 99.40 & 90.40 & 38.00 & 75.90 & \textbf{86.70} & 71.10 & 48.30 & \underline{68.70}  \\
$\text{BAS}$~\cite{wu2021background} & CVPR$22$ & ResNet50 & 28$\times$28 & 99.41 & 95.13 & \underline{71.92} & 88.82 & \underline{84.80} & \underline{71.77} & \underline{49.25} & 68.61 \\
$\text{TS-CAM}$~\cite{gao2021ts} & ICCV$21$ & Deit-S & 14$\times$14 & 98.88 & 87.70 & 49.93 & 78.84 & 82.11 & 67.60 & 44.76 & 64.82\\
$\text{SCM}$~\cite{bai2022weakly} & ECCV$22$ & Deit-S & 14$\times$14 &  \underline{99.64} &  \underline{96.60} & 71.73 & \underline{89.32}  & 83.64  & 68.80  &  45.30 &  65.91 \\
\hline
\rowcolor{mygray}
& &  & & \textbf{99.88} & \textbf{98.45} & \textbf{79.53} & \textbf{92.62}  & 84.42 & \textbf{73.13} & \textbf{56.83} & \textbf{71.46}\\
\rowcolor{mygray}
\multirow{-2}{*}{SAT ({\color{red} ours})}  & \multirow{-2}{*}{{This Work}} & \multirow{-2}{*}{Deit-S} & \multirow{-2}{*}{14$\times$14} & ({\color{red} +0.24}) & ({\color{red} +1.85}) & ({\color{red} +7.61}) & ({\color{red} +3.30}) & ({\color{gray} -2.28}) & ({\color{red} +1.36}) & ({\color{red} +7.58}) & ({\color{red} +2.76}) \\
\hline
\Xhline{2.\arrayrulewidth}
\end{tabular}
\end{center}
\vspace{-2mm}
\caption{\textbf{Localization quality.} Comparison of localization accuracy under different IoU thresholds ($\delta$). Best results are highlighted in \textbf{bold}, second are \underline{underlined}. The last line indicates the improvement (decrease) of our method over the previous best method.} 
\label{table:MaxBoxAccV2}
\end{table*}

\myPara{Implementation Details.}
We evaluate our method on the Deit-S~\cite{touvron2021training} pre-trained on ImageNet~\cite{russakovsky2015imagenet}. On CUB-200~\cite{wah2011caltech}, the training process lasts 30 epochs with a batch size of 256. For ImageNet~\cite{russakovsky2015imagenet}, we train 5 epochs and set batch size to 512. In the training phase, the input images are resized to 256$\times$256 and then randomly cropped to 224$\times$224. In the inference phase, following \cite{wu2021background,guo2021strengthen,zhang2020rethinking}, we adopt ten crop augmentations to obtain classification results and replace random crop with center crop for localization.

\subsection{Ablation Study}
In this subsection, we implement a series of ablation experiments with Deit-S~\cite{touvron2021training} as the backbone. All experiments are conducted on ImageNet~\cite{russakovsky2015imagenet}, a universal dataset, to increase the generalizability of the experimental results.

\myPara{Ablation studies of SAT components.}
Table \ref{table:Ablation} shows the localization accuracy of SAT with different compositions, where we retain the structure of SAT as the baseline. It can be noted that the proposed baseline can effectively activate the object region and already exceeds SOTA method SCM with only $\mathcal{L}_{cls}$. Based on the baseline, the addition of $\mathcal{L}_{ba}$ significantly improves the GT-known Loc with a \textbf{1.78\%} gain, by limiting the expansion of the localization map and forcing it to focus on the regions relevant to the classification. Then the use of $\mathcal{L}_{norm}$ can effectively increase the distinction between the foreground and background regions, resulting in a remarkable increase of \textbf{1.71\%} GT-known Loc. Finally, applying gaussian filter can further eliminate noise and increase local connectivity, by considering the consistency of adjacent patches in the calculation of $\mathcal{L}_{norm}$.

\begin{figure}[t]
\footnotesize
\centering
	\begin{overpic}[width=1\linewidth]{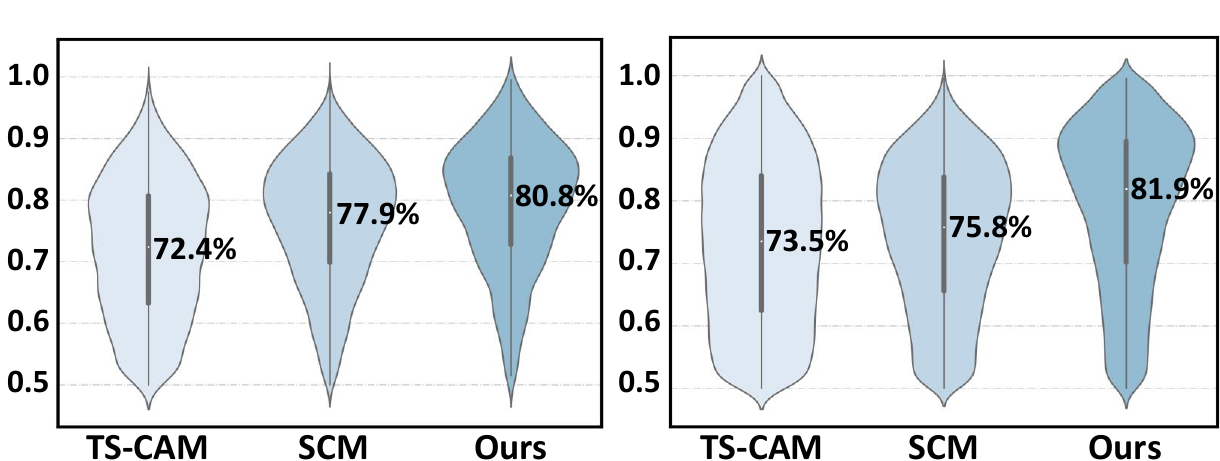}
	    \put(20.5, 35.9){{\textbf{CUB-200}}}
	    \put(70.5, 36){{\textbf{ImageNet}}}
\end{overpic}
\caption{\textbf{Statistical analysis} about localization quality.}
\label{fig:Localization _Quality}
\end{figure}

\myPara{Hyperparameters.}
Fig.~\ref{fig:ablation} presents the sensitivity of localization quality to the hyperparameters $\lambda$ and batch size in $\mathcal{L}_{ba}$ (Eq.~\ref{eq:batch_area}). $\lambda$ denotes the constraint for the average area of the localization maps. A larger $\lambda$ encourages localization map to learn more regions relevant to the classification. Conversely, when $\lambda$ is small, the region learned by the localization map is small but reliable. Fig.~\ref{fig:ablation} (a) shows that the model behaves insensitive to $\lambda$, since GT-known Loc remains stable over a large interval. Batch size affects the calculation of the average area of the localization map within a batch. A sufficiently large batch size can ensure tolerance of area variation between instances. As shown in Fig.~\ref{fig:ablation} (b), when the batch size is larger than 64, the effect of batch on GT-known Loc is limited, with a change of less than 0.5\%.

\subsection{Performance}
\myPara{Main Results.}
Table \ref{table:comparsion} compares SAT with state-of-the-art methods on CUB-200~\cite{wah2011caltech} and ImageNet~\cite{russakovsky2015imagenet}. Experiments show that SAT achieves stable and consistent improvements on both benchmark datasets and significantly outperforms all methods using various backbones. Compared to the CNN counterpart ResNet50-ISIC~\cite{wei2022weakly}, which uses three separate networks, while we use only one network and surpass ResNet50-ISIC by \textbf{1.13\%} and \textbf{3.12\%} GT-known Loc on the CUB-200 and ImageNet, respectively. These quantitative results demonstrate the superiority of the proposed SAT in terms of simplicity and effectiveness. In addition, on CUB-200~\cite{wah2011caltech}, SAT achieves a remarkable performance of \textbf{80.96\%}/\textbf{98.45\%} on Top-1/GT-known Loc, exceeding the baseline method TS-CAM~\cite{gao2021ts} by \textbf{9.66\%}/\textbf{10.75\%}. On ImageNet~\cite{russakovsky2015imagenet}, we surpass the SOTA method SCM~\cite{bai2022weakly} by a large margin, achieving a \textbf{4.33\%} boost in GT-known Loc under the same Deit-S~\cite{touvron2021training} backbone. To further demonstrate the effectiveness of SAT, we visualize the localization maps of our method and TS-CAM in Fig.~\ref{experiment_visual}. In addition to generating sharper and more complete localization maps than TS-CAM, SAT demonstrates robust localization ability in various complex and challenging scenarios. More accuracy results and visualizations on six datasets are provided in the supplementary materials.

\begin{table}[t]
\begin{center}
\small
\renewcommand{\arraystretch}{1.}
\renewcommand{\tabcolsep}{1pt}
\begin{tabular}{l|c|c|cc|cc}
\Xhline{2.\arrayrulewidth}
\hline 
\multirow{2}{*}{\textbf{Method}} & \multirow{2}{*}{\textbf{Venue}}  & \multirow{2}{*}{\textbf{Backbone}} & \multicolumn{2}{c|}{\textbf{CUB-200}}  & \multicolumn{2}{c}{\textbf{OpenImages}} \\
\cline{4-7}
 & & & \textbf{pIoU} & \textbf{PxAP} & \textbf{pIoU} & \textbf{PxAP} \\
\Xhline{2.\arrayrulewidth}
\hline 
$\text{BagCAMs}$~\cite{zhu2022bagging} & ECCV$22$ & IncepV3 & 60.34 & 81.49 & 49.98 & 65.91 \\
$\text{DA-WSOL}$~\cite{zhu2022weakly} & CVPR$22$ & ResNet50 & 56.18 & 74.70 & 49.68 & 65.42  \\
$\text{BAS}$~\cite{wu2021background} & CVPR$22$ & ResNet50 & 71.04 & 89.20 & 50.72 & 66.86 \\
$\text{TS-CAM}$~\cite{gao2021ts} & ICCV$21$ & Deit-S & 66.55 & 81.49 & 43.65 & 54.26 \\
$\text{SCM}$~\cite{bai2022weakly}  & ECCV$22$  & Deit-S & 72.07 & 88.15 &  - & -  \\
\hline
\rowcolor{mygray}
SAT ({\color{red} ours}) & This work & Deit-S & \textbf{77.21} & \textbf{89.87} & \textbf{59.54} & \textbf{74.20} \\
\hline 
\Xhline{2.\arrayrulewidth}
\end{tabular}
\end{center}
\vspace{-2mm}
\caption{\textbf{Segmentation quality} of the localization map compared with other SOTA methods. IncepV3: InceptionV3.}
\label{table:pxap}
\end{table}

\myPara{Localization Quality.}
In Table \ref{table:MaxBoxAccV2}, we compare the localization performance with SOTA methods under the MaxBoxAccV2~\cite{choe2020evaluating} criterion. Despite the limitation of localization map resolution (14$\times$14), we achieve excellent results under different IoU thresholds. Notably, at $\delta$ of 0.7, a strict criterion for the localization quality, the proposed SAT exceeds the previous best method by a significant margin, with \textbf{7.61\%} and \textbf{7.58\%} improvement on CUB-200~\cite{wah2011caltech} and ImageNet~\cite{russakovsky2015imagenet}, respectively. We further plot the IoU distribution of correct bounding boxes in Fig.~\ref{fig:Localization _Quality}, following DANet~\cite{xue2019danet}. Compared to the latest SCM~\cite{bai2022weakly}, with the same Deit-S~\cite{touvron2021training} backbone, we obtain a median IoU gain of \textbf{2.9\%} and \textbf{6.1\%} on CUB-200 and ImageNet. The IoU distribution map demonstrates that the localization results generated by SAT have higher localization quality, especially on ImageNet~\cite{russakovsky2015imagenet}, a universal and challenging dataset.

\begin{figure}[t]
\centering
\begin{overpic}[width=1\linewidth]{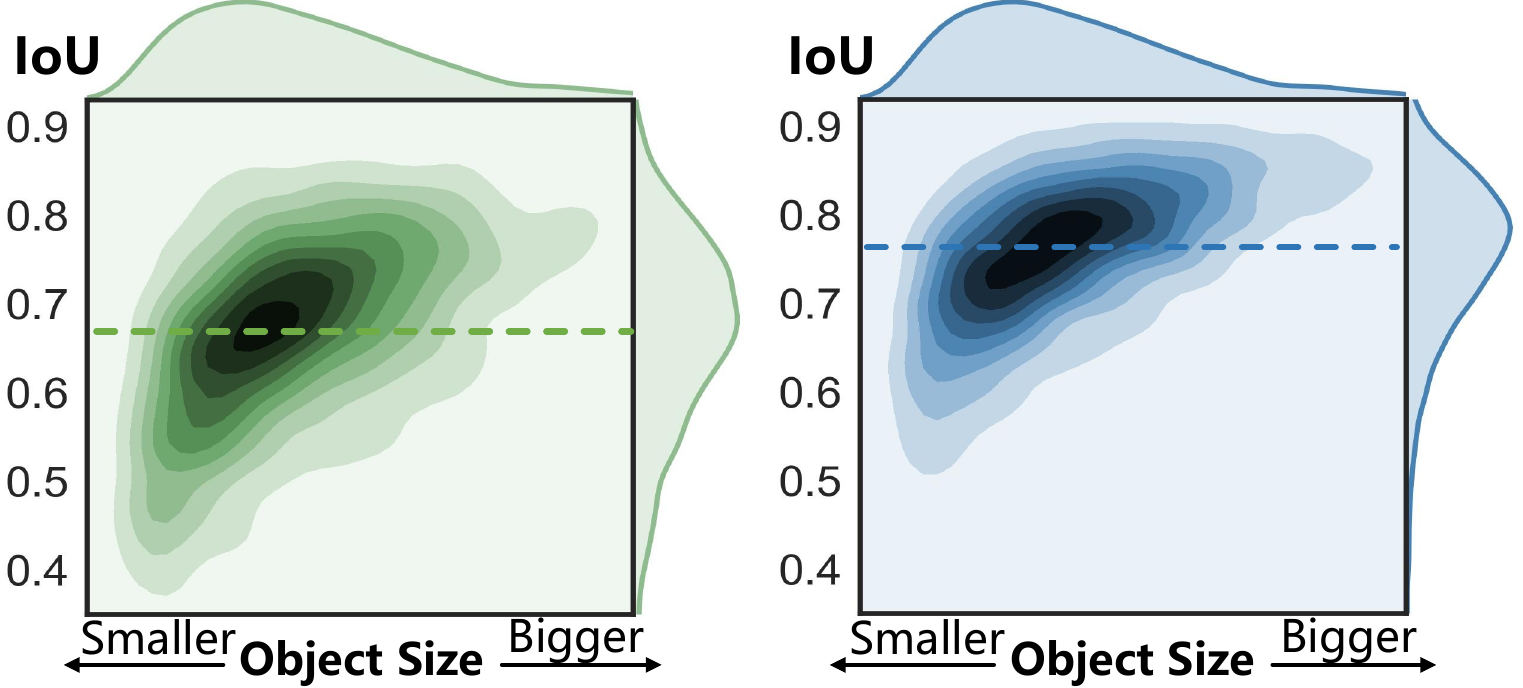}
    
    \put(23, 8){\textbf{TS-CAM}}
    \put(80.5, 8){\textbf{Ours}}
    
\end{overpic}
\caption{\textbf{Density distribution map} about IoU and object size on CUB-200, both measured using pixel-level masks.}
\label{fig:Segmentation_Quality}
\end{figure}

\myPara{Segmentation Quality.}
We evaluate the segmentation performance of SAT on CUB-200~\cite{wah2011caltech} and OpenImages~\cite{choe2020evaluating} since pixel-level labels are available during evaluation. As shown in Table \ref{table:pxap}, the proposed SAT substantially exceeds all methods in pIoU and PxAP metrics. Especially compared to the baseline method TS-CAM~\cite{gao2021ts}, we achieve a remarkable increase of \textbf{10.66\%} and \textbf{15.89\%} pIoU on CUB-200 and OpenImages. To further analyze the segmentation quality, we draw the density distribution map about IoU and object size on CUB-200 in Fig.~\ref{fig:Segmentation_Quality}. It can be observed the proposed SAT achieves an overall improvement and more stable performance on arbitrary sized objects compared to TS-CAM. Nevertheless, there still exists an imbalance segmentation ability of different sized objects. This is mainly due to the limitation of the localization map resolution, which hinders the acquisition of fine segmentation results.

\begin{figure}[t]
\centering
\small
    \begin{overpic}[width=1\linewidth]{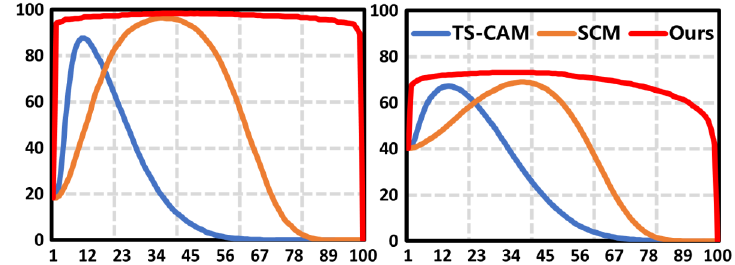}
	    \put(8, 5){\small\textbf{CUB-200}}
	    \put(56.5, 5){\small\textbf{ImageNet}}
	    
	    
	    \put(0, 7){\rotatebox{90}{\small\textbf{GT-known Loc}}} 
    \end{overpic}  
    \caption{\textbf{Threshold sensitivity.} GT-known Loc-Threshold curves for different methods on CUB-200 and ImageNet.}
    \label{fig:Threshold}
\end{figure}

\begin{table}[t] 
\begin{center}
\begin{minipage}{0.65\linewidth}
\renewcommand{\arraystretch}{1}
\renewcommand{\tabcolsep}{1pt} 
\small 
\begin{tabular}{c|cc|cc}
\Xhline{2.\arrayrulewidth}
\hline 
\textbf{Localization} &  \multicolumn{2}{c|}{\textbf{Cls Acc.}} & \multicolumn{2}{c}{\textbf{Loc Acc.}}\\
\textbf{Generation} & \textbf{Top-1} & \textbf{Top-5} & \textbf{Top-1} & \textbf{GT-k.} \\
\Xhline{2.\arrayrulewidth}
\hline 
Attention map & 75.89 & 92.77 & 55.98 & 69.74  \\ 
Spatial token  & \textbf{78.41} & \textbf{94.46}  & \textbf{60.15} & \textbf{73.13}\\  
+$\Delta$  & {\color{red} +2.52} & {\color{red} +1.69} & {\color{red} +4.17} & {\color{red} +3.39}\\
\hline
\Xhline{2.\arrayrulewidth}
\end{tabular}
\end{minipage}
\begin{minipage}{0.34\linewidth}
\begin{overpic}[width=1.\linewidth]{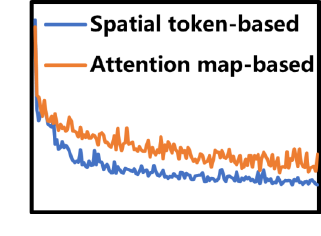}
    \put(36, 0.4){\textbf{\scriptsize{Iterations}}}
    
    \put(1, 15){\textbf{\rotatebox{90}{\scriptsize{Spatial losses}}}} 
\end{overpic}
\end{minipage}
\end{center}
\vspace{-2mm}
\caption{\textbf{Accuracy of different baselines} with the same losses. We follow the idea of TS-CAM to extract the attention maps (before softmax and adding a simgoid function) to synthesize the localization map, and apply the proposed losses to it.}
\label{table:effect_token} 
\end{table}

\myPara{Threshold Sensitivity.}
Fig.~\ref{fig:Threshold} illustrates the curves of GT-known Loc under different thresholds on CUB-200 and ImageNet. Obviously, the curves corresponding to the proposed SAT entirely cover the curves of TS-CAM~\cite{gao2021ts} and SCM~\cite{bai2022weakly}, indicating that SAT performs stable and achieves better results than other methods at arbitrary threshold. Besides, SAT is not sensitive to thresholds and maintains high localization accuracy over a large threshold interval. This phenomenon shows that the localization map generated by SAT has high confidence and excellent visualization.

\myPara{Generalizability.}
To validate the generalizability of the proposed SAT, we implement experiments on different Transformers and various scales. As reported in Table~\ref{table:transformer_backbones}, the proposed approach can be well applied to Deit~\cite{touvron2021training} and Conformer~\cite{peng2021conformer} on CUB-200~\cite{wah2011caltech}. On Deit-T/Deit-B, we achieve \textbf{3.2\%}/\textbf{4.3\%} GT-known Loc improvement compared to SCM~\cite{bai2022weakly}, and \textbf{3.3\%}/\textbf{2.1\%} Top-1 classification accuracy gain over TS-CAM~\cite{gao2021ts}. Besides, SAT also achieves state-of-the-art results on Conformer-S. Quantitative experiments show that the proposed SAT achieves excellent performance in both tasks on various transformer backbones.

\begin{table}[t]
\begin{center}
\renewcommand{\arraystretch}{1}
\renewcommand{\tabcolsep}{1.4pt}
\small
\begin{tabular}{c|cccc|cccc}
\Xhline{2.\arrayrulewidth}
\hline 
\multirow{2}{*}{\textbf{Model}}   & \multicolumn{4}{c|}{\textbf{Top-1 Cls}} & \multicolumn{4}{c}{\textbf{GT-known Loc}}\\
& TS-CAM & SCM & \multicolumn{2}{c|}{Ours (+$\Delta$)}& TS-CAM & SCM & \multicolumn{2}{c}{Ours (+$\Delta$)}\\
\hline
Deit-T & 72.9 & - & \multicolumn{2}{c|}{\cellcolor{mygray}\textbf{76.2} ({\color{red} +3.3})}& 86.4 & 91.8 & \multicolumn{2}{c}{\cellcolor{mygray}\textbf{95.0} ({\color{red} +3.2})}\\
Deit-S & 80.3 & 77.1 & \multicolumn{2}{c|}{\cellcolor{mygray}\textbf{82.1} ({\color{red} +1.8})} & 87.7 &  96.6 & \multicolumn{2}{c}{\cellcolor{mygray}\textbf{98.5} ({\color{red} +1.9})} \\
Deit-B & 83.2 &  - & \multicolumn{2}{c|}{\cellcolor{mygray}\textbf{85.3} ({\color{red} +2.1})} & 83.3 &  93.8 & \multicolumn{2}{c}{\cellcolor{mygray}\textbf{98.1} ({\color{red} +4.3})} \\
\hline
Conf-S & 81.0 &  - & \multicolumn{2}{c|}{\cellcolor{mygray}\textbf{82.4} ({\color{red} +1.4})}  & 94.1 &  96.1 & \multicolumn{2}{c}{\cellcolor{mygray}\textbf{97.9} ({\color{red} +1.8})} \\
\hline
\Xhline{2.\arrayrulewidth}
\end{tabular}
\end{center} 
\vspace{-2mm}
\caption{\textbf{Generalizability} on CUB-200. $\Delta$ denotes the improvement of SAT over the previous best method. Conf-S: Conformer-S.}
\label{table:transformer_backbones}
\end{table}

\subsection{Discussions}

\begin{table}[t]
    \begin{center}
    \begin{minipage}{0.54\linewidth} 
    \renewcommand{\arraystretch}{1.}
    \renewcommand{\tabcolsep}{1.2pt}
    \small
    \begin{tabular}{c|l|ccc}
    \Xhline{2.\arrayrulewidth}
    \hline  
    &\multirow{2}{*}{\textbf{Method}} &\multicolumn{3}{c}{\textbf{Loc Acc.}}\\
     & & \textbf{Top-1} & \textbf{Top-5} & \textbf{GT-k.}\\
    \Xhline{2.\arrayrulewidth}
    \hline 
    \textbf{(a)} & ORNet & 73.39 & 84.31 & 88.38 \\
    \textbf{(b)} & BAS & 49.59 & 60.68 & 63.76 \\
    \textbf{(c)} & BAS* & 75.18 & 88.66 & 92.87 \\
    \hline
    \rowcolor{mygray}
    \textbf{(d)} & Ours &\textbf{80.96} & \textbf{94.13} & \textbf{98.45}\\
    \hline
    \Xhline{2.\arrayrulewidth}
    \end{tabular} 
    \end{minipage}
    \begin{minipage}{0.45\linewidth}
    \centering
    \begin{overpic}[width=1.\linewidth]{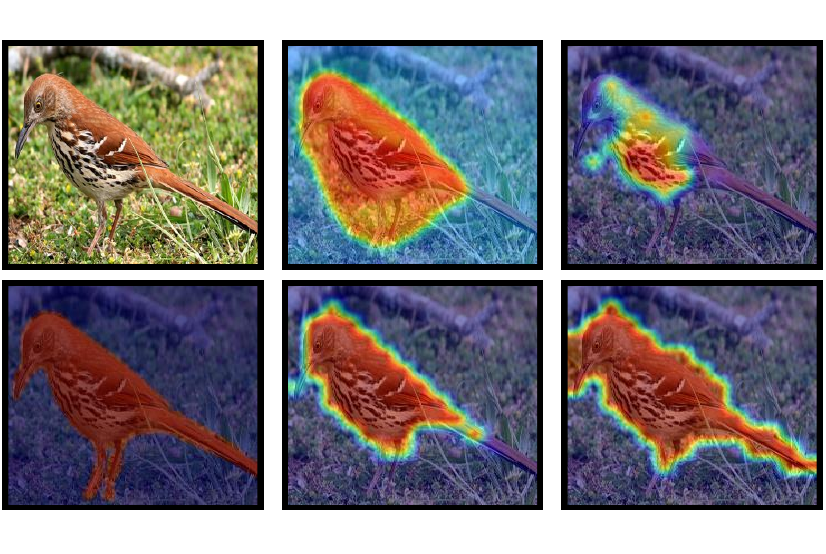}
        \put(8, 64){\textbf{\scriptsize{Image}}}
        \put(9, 0){\textbf{\scriptsize{label}}}
        \put(46, 0){\textbf{\scriptsize{(c)}}}
	    \put(80, 0){\textbf{\scriptsize{(d)}}}
	    \put(46, 64){\textbf{\scriptsize{(a)}}}
	    \put(80, 64){\textbf{\scriptsize{(b)}}}
    \end{overpic}
    
    \end{minipage}
    \end{center}
    \vspace{-2mm}
    \caption{\textbf{SAT $vs$ FPM-based methods} on Deit-S backbone. For ORNet, we replace the classification network with Deit-S. For BAS, we insert the generator in the 10-th self-attention module. * denotes replacing the original losses with our proposed losses.}
    \label{table:fpm}
\end{table}

\myPara{Why spatial token-based structure is critical?}
To clarify this question, we follow TS-CAM~\cite{gao2021ts} to construct a attention map-based baseline and apply the proposed losses on it for comparison. As shown in Table~\ref{table:effect_token}, the attention map-based method substantially reduces the classification and localization accuracy, with \textbf{2.52\%} and \textbf{3.39\%} decrease in Top-1 Cls and GT-known Loc, respectively. To explore how this structural difference affects the accuracy, we plot the curves of spatial losses during training in Table~\ref{table:effect_token}. It can be noticed that the potential optimization conflicts of attention map-based method will lead to poor convergence of losses and insufficient learning, thus limiting the model to benefit from the losses. Experimental results show that the proposed spatial token-based structure can release the potential of the model by more efficient localization learning and avoiding optimization conflicts between two tasks, which are essential for the accuracy improvement.

To further verify the effectiveness of the proposed structure in generating localization map, we reproduce two FPM-based methods ORNet~\cite{xie2021online} and BAS~\cite{wu2021background} on the Deit-S. Both of them produce localization maps by a generator. As shown in Table~\ref{table:fpm}, first, we note that the original BAS is not suitable for the transformer due to the skip-connection in transformer causing inaccurate evaluation of the masking effect. Second, both FPM-based methods suffer from incomplete localization problem. This may be because they can only mask the input of a specific layer, resulting in overfitting the features that the specific feature layer focuses on. While the proposed structure can extract localization knowledge from different layers in an efficient form, thus improving localization ability and robustness. Quantitative and qualitative results demonstrate the effectiveness of SAT in terms of localization map generation and learning.

\begin{figure}[t] 
\centering
    \begin{overpic}[width=1\linewidth]{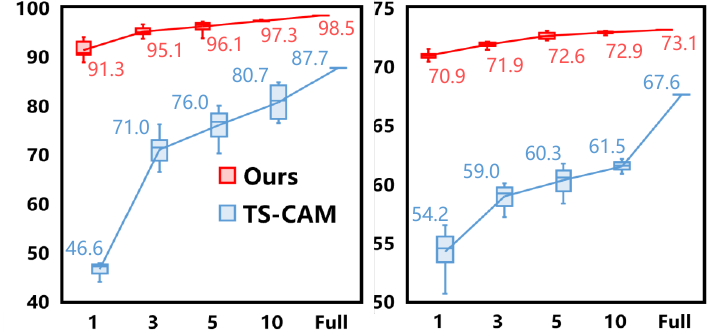}
	\put(30.2, 7.){\textbf{CUB-200}}
	\put(77.1, 7.){\textbf{ImageNet}}

    \put(-0.3, 12){\rotatebox{90}{\small\textbf{GT-known Loc}}}     
    \end{overpic} 
    \caption{\textbf{Few-shot learning.} Box plot of the k-shot localization performance. For each k-shot setting, we repeat the experiment 10 times to randomly select k images as training data, and record the experimental results. The average results are shown in the figure.}
    \label{fig:few_shot}
\end{figure}

\myPara{Data-efficiency and tuning-efficiency.} To validate the data-efficiency of the proposed method, we evaluate the localization performance of SAT and TS-CAM~\cite{gao2021ts} under few-shot setting in Fig.~\ref{fig:few_shot}. Compared to TS-CAM, SAT achieves promising results with only a few training data. Even in the extreme setting of only 1 training image per class, we exceed TS-CAM trained with full data by \textbf{3.6\%} and \textbf{3.3\%} GT-known Loc on CUB-200 and ImageNet, respectively. The box plot verifies the effectiveness and efficiency of the proposed SAT that implements localization task by an extra lightweight spatial token and enables high degree of network parameter sharing, thus requiring only a few data to achieve excellent localization results. In contrast, TS-CAM requires large amounts of data to fine-tune the classification feature maps to serve the localization task.

To explore the tuning-efficiency of SAT, we freeze different parts of the network and evaluate the GT-known Loc of the proposed method under few-shot setting. All experiments are performed with ImageNet pre-trained weights. As shown in Table~\ref{table:trade-off}, SAT can still obtain competitive results when freezing most parameters of the pre-trained classification model. It can be analyzed that the task-specific spatial token can be well adapted to the classification features and plays a major role in the localization learning, which allows efficient implementation of the localization task with fewer parameters to be fine-tuned. Experiments demonstrate both the data-efficiency and the tuning-efficiency of SAT. When freezing \textbf{81\%} of the parameters and using only 1 image per class, SAT exceeds the SOTA method SCM~\cite{bai2022weakly} by \textbf{2.1\%} GT-known Loc on ImageNet.

\begin{table}[]
\small
\begin{center}
\renewcommand{\arraystretch}{1.}
\renewcommand{\tabcolsep}{1.2pt}
\begin{tabular}{c|c|ccccc|c|ccccc}
\Xhline{2.\arrayrulewidth}
\hline 
\multirow{2}{*}{} & \multirow{2}{*}{\makecell{Fro.\\rate}} & \multicolumn{5}{c|}{\textbf{CUB-200}~\cite{wah2011caltech}}          & \multirow{2}{*}{\makecell{Fro.\\rate}} & \multicolumn{5}{c}{\textbf{ImageNet}~\cite{russakovsky2015imagenet}}     \\

&   & 1    & 3    & 5    & 10   & Full & & 1    & 3    & 5    & 10   & Full \\
\Xhline{2.\arrayrulewidth}
\hline 
\textbf{(a)} & 0\%       & \cellcolor{orange!11}91.3& \cellcolor{orange!47}95.1& \cellcolor{orange!58}96.1& \cellcolor{orange!72}97.3& \cellcolor{orange!85}98.5& 0\%         & \cellcolor{orange!11}70.9& \cellcolor{orange!53}71.9& \cellcolor{orange!72}72.6& \cellcolor{orange!80}72.9& \cellcolor{orange!85}73.1\\
\textbf{(b)} & 24\%      & \cellcolor{orange!29}93.5& \cellcolor{orange!56}95.9& \cellcolor{orange!66}96.8& \cellcolor{orange!74}97.5& \cellcolor{orange!81}98.1& 21\%        & \cellcolor{orange!41}71.6& \cellcolor{orange!63}72.3& \cellcolor{orange!71}72.6& \cellcolor{orange!77}72.8& \cellcolor{orange!85}73.1\\
\textbf{(c)} & 48\%      & \cellcolor{orange!61}96.4& \cellcolor{orange!65}96.7& \cellcolor{orange!68}97.0& \cellcolor{orange!73}97.4& \cellcolor{orange!79}98.0        & 42\%        & \cellcolor{orange!50}71.8& \cellcolor{orange!58}72.1& \cellcolor{orange!66}72.4& \cellcolor{orange!74}72.7& \cellcolor{orange!80}72.9\\
\textbf{(d)} & 71\%      &\cellcolor{orange!32}93.8& \cellcolor{orange!41}94.6& \cellcolor{orange!45}94.9& \cellcolor{orange!48}95.2& \cellcolor{orange!56}95.9& 64\%        & \cellcolor{orange!28}71.3& \cellcolor{orange!33}71.4& \cellcolor{orange!46}71.7& \cellcolor{orange!53}71.9& \cellcolor{orange!61}72.2\\
\textbf{(e)} & 91\%      & \cellcolor{orange!13}92.1& \cellcolor{orange!18}92.5& \cellcolor{orange!21}92.8& \cellcolor{orange!24}93.1& \cellcolor{orange!28}93.4& 81\%        & \cellcolor{orange!11}70.9& \cellcolor{orange!20}71.1& \cellcolor{orange!24}71.2& \cellcolor{orange!28}71.3& \cellcolor{orange!32}71.4\\
\Xhline{2.\arrayrulewidth}
\hline

\end{tabular}
\end{center}
\vspace{-2mm}
\caption{\textbf{Frozen rate $w.r.t.$ few-shot learning.} The frozen parts are as follows: \textbf{(a)} None. \textbf{(b)} Attention layer of transformer blocks. \textbf{(c)} MLP layer of transformer blocks. \textbf{(d)} Transformer blocks. \textbf{(e)} Position embedding, projection, transformer blocks, and MLP layer of spatial aware transformer blocks. Fro.: Frozen.}
\label{table:trade-off}
\end{table}


\begin{figure}[t]
\centering
\small 
	\begin{overpic}[width=1\linewidth]{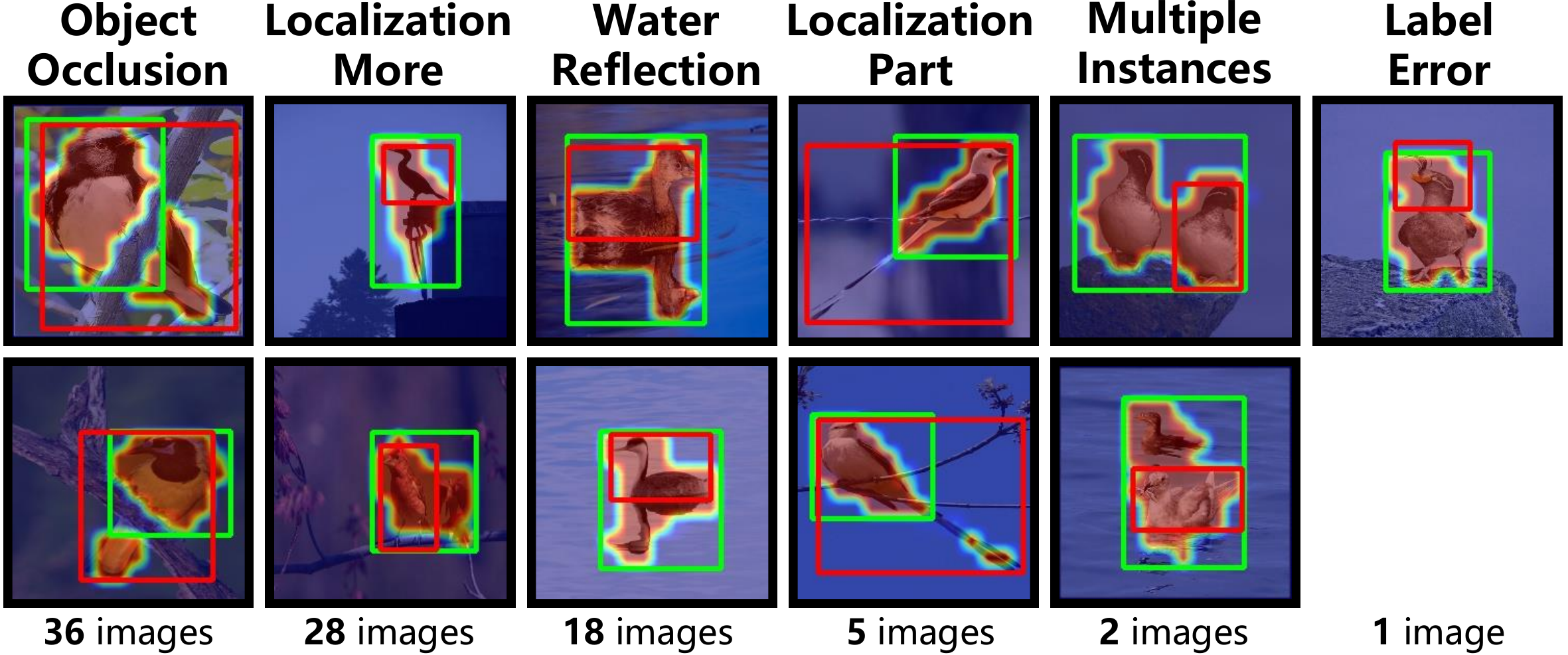}
\end{overpic}
\caption{\textbf{Localization error analysis} on CUB-200.}
\label{fig:error}
\end{figure}

\subsection{Limitations} 
As shown in Fig.~\ref{fig:error}, we count all localization errors (90 images) on CUB-200 test set (5,794 images) and classify the error causes into six categories, including object occlusion, localization more, water reflection, localization part, multiple instances, and label error. Among them, object occlusion causes the object to be split into two or more parts, resulting in incomplete localization results. Localization more is usually due to the positive effect of co-occurrence context on classification, leading
to localizing confounding background regions. Water reflection is an inherent problem of WSOL and difficult to be solved with only image-level labels. Therefore, future work needs to consider more the interaction between object and context to overcome the problems of object occlusion and localization more.

\section{Conclusion}
This paper proposes to introduce a spatial-aware token (SAT) specific to the localization task, instead of synthesizing the classification feature map, thus avoiding optimization conflicts
between classification and localization tasks. To this end, we construct a spatial-query attention (SQA) module to generate localization map and extract localization knowledge from the classification task. Meanwhile, two spatial constraints, including batch area loss and regularization loss, are designed to compensate for the insufficient supervision. Extensive experiments on multiple benchmarks verify the effectiveness and efficiency of the proposed SAT that surpasses previous methods by a large margin.

\textbf{Acknowledgments.} 
This work is supported by National Key R\&D Program of China under Grant 2020AAA0105700, National Natural Science Foundation of China (NSFC) under Grants 62225207, U19B2038 and 62121002.

{\small
\bibliographystyle{IEEEtran}
\bibliography{main.bib}
}

\clearpage

\appendix

\section*{Supplementary Materials}

\section{Ablation Study}


\begin{table*}[b] 
\begin{center}
\renewcommand{\arraystretch}{1.}
\renewcommand{\tabcolsep}{5.pt}
\begin{tabular}{c|ccc|ccc}
\Xhline{2.\arrayrulewidth}
\hline 
\multirow{2}{*}{\textbf{N}} & \multicolumn{3}{c|}{\textbf{CUB-200}} & \multicolumn{3}{c}{\textbf{ImageNet}}\\

 & \textbf{Top-1 Cls} &
 \textbf{Top-1 Loc} & \textbf{GT-k. Loc} & \textbf{Top-1 Cls} &
 \textbf{Top-1 Loc} & \textbf{GT-k. Loc}\\
\Xhline{2.\arrayrulewidth}
\hline 
1 & 81.45 & 79.82 & 97.48 & 78.16  & 59.04 & 71.84  \\
2 & 81.62 & 80.17 &  98.02 & 78.33 & 59.90 & 72.77 \\
\rowcolor{mygray}
3 & \textbf{82.05} & \textbf{80.96} & \textbf{98.45} & \textbf{78.41} & \textbf{60.15} & \textbf{73.13}  \\
4 & 81.93 & 80.76 & 98.36 & 78.23 & 59.87 & 72.92  \\
5 & 80.69 & 78.75 & 97.12 & 78.16 & 58.67 & 71.41 \\
\hline
\Xhline{2.\arrayrulewidth}
\end{tabular}
\end{center}
\vspace{-2mm}
\caption{\textbf{Number of spatial aware transformer blocks.} We select the 10-th block as the spatial aware transformer block when N = 1, and the 10-th and 11-th blocks when N=2. When N $>$ 2, the last N blocks are adopted as spatial aware transformer blocks.}
\label{table:N}
\end{table*}

\begin{table*}[b] 
\begin{center}
\renewcommand{\arraystretch}{1.}
\renewcommand{\tabcolsep}{5.pt}
\begin{tabular}{c|ccc|ccc}
\Xhline{2.\arrayrulewidth}
\hline 
\multirow{2}{*}{\textbf{Dot Position}} & \multicolumn{3}{c|}{\textbf{CUB-200}} & \multicolumn{3}{c}{\textbf{ImageNet}}\\

 & \textbf{Top-1 Cls} & \textbf{Top-1 Loc} & \textbf{GT-k. Loc} & \textbf{Top-1 Cls} &
 \textbf{Top-1 Loc}& \textbf{GT-k. Loc}\\
\Xhline{2.\arrayrulewidth}
\hline 
Before softmax & 81.67 & 80.39
 & 98.19 & 77.78 &  59.57 & 72.85 \\
\rowcolor{mygray}
After softmax & \textbf{82.05} & \textbf{80.96} & \textbf{98.45} & \textbf{78.41} & \textbf{60.15} & \textbf{73.13}  \\
\hline
\Xhline{2.\arrayrulewidth}
\end{tabular}
\end{center}
\vspace{-2mm}
\caption{\textbf{Dot position.} Ablation experiments on the position of the dot product in the spatial-query
 attention module.}
\label{table:dot_position}
\end{table*}

\begin{table*}[t] 
\begin{center}
\renewcommand{\arraystretch}{1.}
\renewcommand{\tabcolsep}{5.pt}
\begin{tabular}{c|cc|cc|ccc}
\Xhline{2.\arrayrulewidth}
\hline 
& \multicolumn{2}{c|}{\textbf{Initial Weights}} & \multicolumn{2}{c|}{\textbf{Cls Acc.}} & \multicolumn{3}{c}{\textbf{Loc Acc.}}\\

& \textbf{Class token} & \textbf{Spatial token} & \textbf{Top-1} & \textbf{Top-5} & \textbf{Top-1} & \textbf{Top-5} &  \textbf{GT-k.}\\
\Xhline{2.\arrayrulewidth}
\hline 
(a) & \multicolumn{2}{c|}{Pre-trained (shared)} & 77.83 & 93.92 & 58.31 & 68.35 & 71.08  \\
(b) & Pre-trained &  Pre-trained & 78.34
 & 94.14 & 59.81 & 69.96 & 72.60  \\
\rowcolor{mygray}
(c)  & Pre-trained &  Random initial & \textbf{78.41} & \textbf{94.46} & \textbf{60.15} & \textbf{70.52} & \textbf{73.13}   \\
\hline
\Xhline{2.\arrayrulewidth}
\end{tabular}
\end{center}
\vspace{-2mm}
\caption{\textbf{Class token $w.r.t.$ spatial-aware token.} (a) Class token and spatial token share the same token, and its initial weights are the pre-trained weights of class token. (b) Class token and spatial token use separate tokens, and their initial weights are the pre-trained weights of class token. (c) Class token and spatial token use separate tokens, where the initial weights of the spatial token are randomly initialized.}
\label{table:exploratory}
\end{table*}

\myPara{Number of spatial aware transformer blocks.}
We fix the whole network to 12 blocks and adjust the number of spatial aware transformer blocks, denoted as N. As shown in Table~\ref{table:N}, the best results are achieved when N is set to 3, which indicates that fusing the localization maps $M^{l}$ learned from different blocks is helpful to obtain a complete localization result. However, when N is too large, it increases the optimization difficulty of the normalization loss thus reducing the localization performance.

\myPara{Dot position.}
We explore the impact of the position of the dot product in the spatial-query attention module, as shown in Table~\ref{table:dot_position}. Quantitative experiments show that performing the dot product before softmax will reduce the performance of classification and localization, mainly because the exponential form in softmax makes the semantic prediction $M^{l}$ learning insufficient. Therefore, the dot product after the softmax function enables semantic prediction $M^{l}$ to better capture the localization knowledge from the self-attention mechanism.

\section{Analysis}
\myPara{Class token $w.r.t.$ spatial-aware token.}
To analyze the differences between spatial-aware token and class token, we implement the exploratory experiment as shown in Table~\ref{table:exploratory}. From Table~\ref{table:exploratory} (a), it can be analyzed that sharing the same token between class token and spatial token will bring optimization conflict between classification and localization tasks, thus decreasing both classification and localization performance. In addition, as illustrated in Table~\ref{table:exploratory} (b), using separate tokens and initializing the weights of the spatial token to the pre-trained weights of class token also results in reduced localization accuracy, suggesting that the information learned by the spatial token and class token are significantly different. As a result, it is necessary to learn a separate spatial token from scratch.

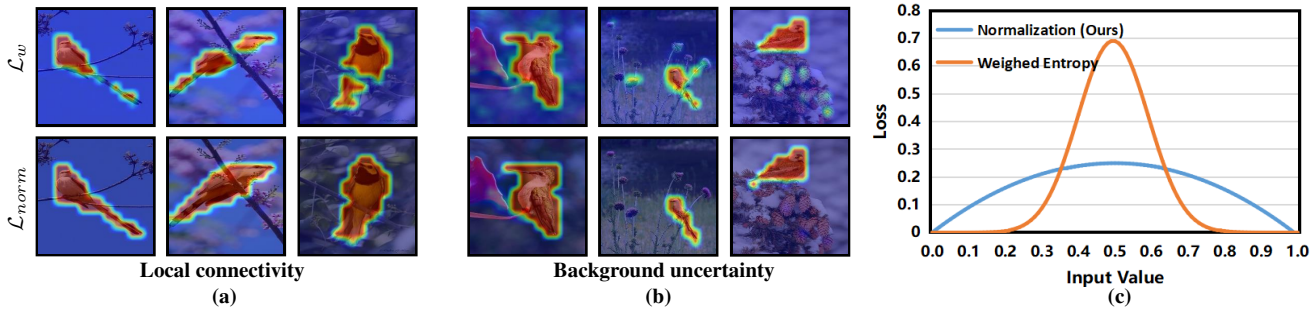
\begin{figure*}[h]
\footnotesize
\centering
	\begin{overpic}[width=1\linewidth]{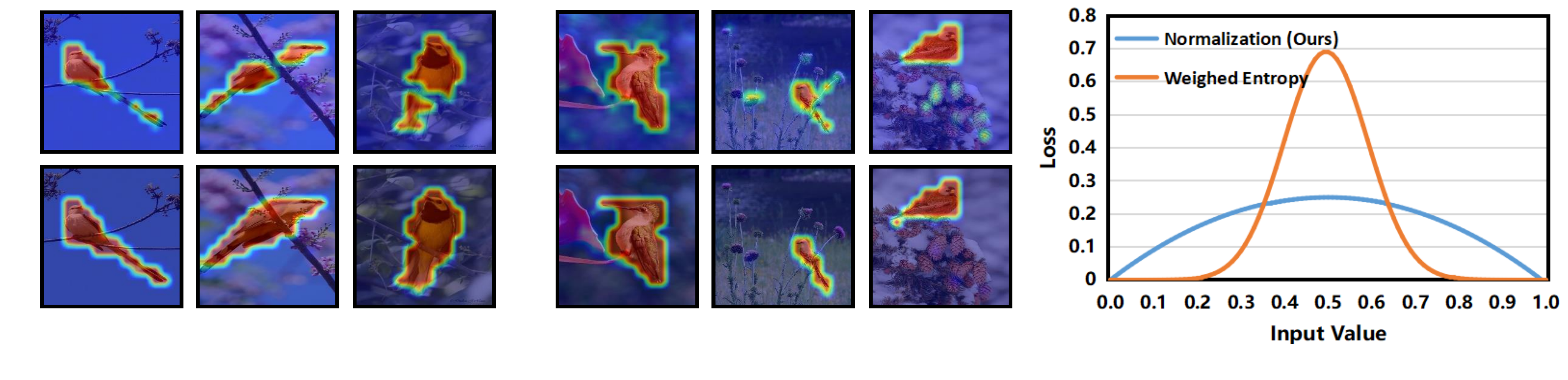} 
        \put(0.5, 6.1){\rotatebox{90}{ $\mathcal{L}_{norm}$}}
        \put(0.5, 18.1){\rotatebox{90}{ $\mathcal{L}_{w}$}}
	    \put(10.5, 3){{\textbf{Local connectivity}}} 
	    \put(42, 3){{\textbf{Background uncertainty}}}
        \put(16, 1){{\textbf{(a)}}}
        \put(49, 1){{\textbf{(b)}}}
        \put(84.2, 1){{\textbf{(c)}}}
	    
\end{overpic}
\caption{\textbf{Comparison between $\mathcal{L}_{w}$ and $\mathcal{L}_{norm}$.} (a) Visual comparison of SAT w/ $\mathcal{L}_{w}$ and SAT w/ $\mathcal{L}_{norm}$ on local connectivity. (b) Visual comparison of SAT w/ $\mathcal{L}_{w}$ and SAT w/ $\mathcal{L}_{norm}$ on background uncertainty. (c) Loss-Input value curves for different loss functions.}
\label{norm_fig}
\end{figure*}

\myPara{Normalization loss $w.r.t.$ weighed entropy loss.}
We compare the weighed entropy loss ($\mathcal{L}_{w}$) in ORNet~\cite{xie2021online} with our proposed normalization loss ($\mathcal{L}_{norm}$) in Table~\ref{table:norm}. Both losses aim to provide pixel-level supervision to increase the distinction between foreground and background of the localization map, but the effects are somewhat different, as shown in Fig.~\ref{norm_fig}. 1) The proposed $\mathcal{L}_{norm}$ includes a gaussian filtering operation to incorporate the values of adjacent patches in the calculation of the loss, thus encouraging the local continuity of the localization map. As illustrated in Fig.~\ref{norm_fig} (a), the localization map generated by SAT w/ $\mathcal{L}_{norm}$ has better connectivity compared to SAT w/ $\mathcal{L}_{w}$. 2) Fig.~\ref{norm_fig} (c) shows the loss curves of the two loss functions versus the input values. Compared to $\mathcal{L}_{norm}$, $\mathcal{L}_{w}$ is already close to zero at input values of 0.2 or 0.8, which indicates that $\mathcal{L}_{w}$ allows the background to be activated with low response, as presented in Fig.~\ref{norm_fig} (b). While using a higher visualization threshold to filter out the background region will reduce the connectivity of localization map generated by SAT w/ $\mathcal{L}_{w}$ in Fig.~\ref{norm_fig} (a), resulting in decreased localization performance. Therefore, $\mathcal{L}_{norm}$ is more suitable for the proposed SAT and SAT w/ $\mathcal{L}_{norm}$ achieves the best results in Table~\ref{table:norm}.

\begin{table}[h] 
\begin{center}
\renewcommand{\arraystretch}{1.}
\renewcommand{\tabcolsep}{3.pt}
\begin{tabular}{c|l|cc|ccc}
\Xhline{2.\arrayrulewidth}
\hline 
& \multirow{2}{*}{\textbf{Method}} & \multicolumn{2}{c|}{\textbf{Cls Acc.}} & \multicolumn{3}{c}{\textbf{Loc Acc.}}\\

&   & \textbf{Top-1} & \textbf{Top-5} & \textbf{Top-1} & \textbf{Top-5} &  \textbf{GT-k.}\\
\Xhline{2.\arrayrulewidth}
\hline 
(a) & SAT w/ $\mathcal{L}_{w}$ & 81.64 & 95.22 & 78.68 & 91.82 & 96.27  \\ 
\rowcolor{mygray}
(b) & SAT w/ $\mathcal{L}_{norm}$ & \textbf{82.05} & \textbf{95.56} & \textbf{80.96} & \textbf{94.13} & \textbf{98.45}   \\
\hline
\Xhline{2.\arrayrulewidth}
\end{tabular}
\end{center}
\vspace{-2mm}
\caption{\textbf{Normalization loss $w.r.t.$ weighed entropy loss.} The accuracy of our method using normalization loss $\mathcal{L}_{norm}$ and weighed entropy $\mathcal{L}_{w}$ loss on CUB-200, respectively.}
\label{table:norm}
\end{table}

\begin{table*}[h] 
\begin{center}
\begin{minipage}{0.61\linewidth}
\renewcommand{\arraystretch}{1}
\renewcommand{\tabcolsep}{4pt}  
\begin{tabular}{c|l|c|cc|ccc}
\Xhline{2.\arrayrulewidth}
\hline 
& \multirow{2}{*}{\textbf{Method}} & \multirow{2}{*}{\textbf{Stage}} &  \multicolumn{2}{c|}{\textbf{Cls Acc.}} & \multicolumn{3}{c}{\textbf{Loc Acc.}}\\

& &  & \textbf{Top-1} & \textbf{Top-5} & \textbf{Top-1} & \textbf{Top-5} &  \textbf{GT-k.}\\
\Xhline{2.\arrayrulewidth}
\hline 
(a) & SAT w/ $\mathcal{L}_{area}$ & one-stage & 79.39 & 95.41 & 25.68 & 32.83 & 34.92  \\ 
(b) & SAT w/ $\mathcal{L}_{area}$ & two-stage & 80.19 & 94.51 & 57.01 & 66.79 & 70.54  \\
\rowcolor{mygray}
(c) & SAT w/ $\mathcal{L}_{ba}$ & one-stage & \textbf{82.05} & \textbf{95.56} & \textbf{80.96} & \textbf{94.13} & \textbf{98.45}   \\
\hline
\Xhline{2.\arrayrulewidth}
\end{tabular}
\end{minipage}
\begin{minipage}{0.38\linewidth}
\begin{overpic}[width=1.\linewidth]{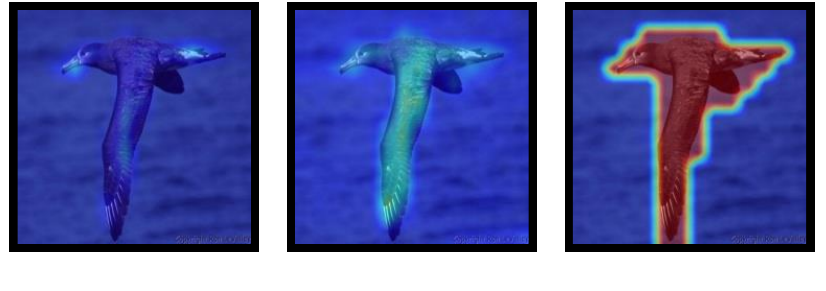}
    \put(14, 0.4){\textbf{\scriptsize{(a)}}}
    \put(47, 0.4){\textbf{\scriptsize{(b)}}}
    \put(80, 0.4){\textbf{\scriptsize{(c)}}}
     
\end{overpic}
\end{minipage}
\end{center}
\vspace{-2mm}
\caption{\textbf{batch area loss $w.r.t.$ area loss.} One-stage indicates training the model in an end-to-end manner. Two-stage means first training the network with classification losses only. Then the weights of the backbone are fixed and only the spatial token is trained with all losses.}
\label{table:ba}
\end{table*}

\myPara{Batch area loss $w.r.t.$ area loss.} 
In Table~\ref{table:ba}, we compare the proposed batch area loss $\mathcal{L}_{ba}$ with the area loss $\mathcal{L}_{area}$ in FPM-based~\cite{meng2021foreground,wu2021background,xie2021online} on CUB-200. Experiments show that $\mathcal{L}_{area}$ cannot be well applied to SAT either in one-stage or two-stage. This is because the generation and learning of the localization map in the SAT occur in the attention module, while in transformer, the token sequence input to the attention module can be propagated to the next layer by the skip-connection, which makes the area loss not suitable for SAT. For this reason, we propose batch area loss, which not only provides a sparse area supervision for the localization maps, but also guarantees the tolerance of area variation between instances. The visualization results and accuracy verify the effectiveness of the proposed batch area loss.

\myPara{Error analysis.}
To further analyze the effect of the proposed method, we count all the localization errors (90 images) on CUB-200~\cite{wah2011caltech} test set (5,794 images) and classify them according to the error causes. As listed in Table~\ref{table:error}, we classify the error causes into the following six categories, including \textbf{object occlusion} (36 images), \textbf{localization more} (28 images), \textbf{water reflection} (18 images), \textbf{localization part} (5 images), \textbf{multiple instances} (2 images), \textbf{label error} (1 image). Specifically, object occlusion causes the object to be split into two or more parts, resulting in incomplete localization results, as shown in Table~\ref{table:error}. Localization more is often due to the positive effect of co-occurrence
context on the classification network, leading to localizing confounding background regions. In addition, water reflection is an inherent challenge for weakly supervised object localization, and it is difficult to achieve correct localization results with only image-level labels. In this way, to achieve better localization performance, future work needs to take more into account the interaction between objects and background to overcome the problems of object occlusion and localization more.

\section{Performance}

\myPara{Tunable parameters.}
We detail the tunable parameters for freezing different parts in Table~\ref{table:parameter}, where we follow the freezing settings of Table 8 in the main text. When freezing 81\% of the parameters, only 1.4M parameters are tunable on the backbone network, which is \textbf{6\%} of the parameters in the entire backbone network (21.7M). In this case, SAT still exceeds the existing transformer-based approaches in both classification and localization with only 4.8M tunable parameters , which verifies the efficiency and effectiveness of the proposed method.

\begin{table*}[t] 
\begin{center}
\renewcommand{\arraystretch}{1.}
\renewcommand{\tabcolsep}{5.pt}
\begin{tabular}{l|c|cc|cc|ccc}
\Xhline{2.\arrayrulewidth}
\hline 
\multirow{2}{*}{\textbf{Methods}} & \multirow{2}{*}{\textbf{Frozen Rate}} & \multicolumn{2}{c|}{\textbf{Tunable Parameters}} &  \multicolumn{2}{c|}{\textbf{Inference}} & \multicolumn{3}{c}{\textbf{Accuracy}}\\

&   & \textbf{Backbone} & \textbf{Head} & \textbf{FLOPs} & \textbf{Parameters} & \textbf{Top-1 Cls} &   \textbf{Top-1 Loc} & \textbf{GT-k. Loc}\\
\Xhline{2.\arrayrulewidth}
\hline 
$\text{TS-CAM}$~\cite{gao2021ts}  & 0\%  & 21.7M & 3.4M & 4.9G & 25.1M & 74.30 & 53.40 & 67.60  \\
$\text{LCTR}$~\cite{chen2021lctr}  & 0\%  & 21.7M & 15.1M &  7.2G & 36.8M & 77.10 & 56.10 & 68.70  \\
$\text{SCM}$~\cite{bai2022weakly}  & 0\%  & 21.7M & 3.4M & 4.9G  & 25.1M & 76.70 & 56.10 & 68.80 \\
\hline
SAT (e)  & 81\%  & \textbf{1.4M} & 3.4M &  4.9G & 25.1M & 77.79 & 58.29 &  71.14 \\
SAT (d)  & 64\%  & 5.8M & 3.4M &  4.9G & 25.1M & 78.23 & 59.37 &  72.20 \\
SAT (c)  & 42\%  & 11.1M & 3.4M &  4.9G &  25.1M & 78.24 & 59.97 &  72.88 \\
SAT (b)  & 21\%  & 16.4M & 3.4M &  4.9G & 25.1M  & 78.12 & 60.00 & 73.10 \\
\rowcolor{mygray}
SAT (a) & 0\%  & 21.7M & 3.4M & 4.9G & 25.1M & \textbf{78.41} & \textbf{60.15} & \textbf{73.13} \\
\hline
\Xhline{2.\arrayrulewidth}
\end{tabular}
\end{center}
\vspace{-2mm}
\caption{\textbf{Tunable parameters.} The frozen parts are as follows: \textbf{(a)} None. \textbf{(b)} Attention layer of transformer blocks. \textbf{(c)} MLP layer of transformer blocks. \textbf{(d)} Transformer blocks. \textbf{(e)} Position embedding, projection, transformer blocks, and MLP layer of spatial aware transformer blocks.}
\label{table:parameter}
\end{table*}
\begin{table*}[t] 
\begin{center}
\renewcommand{\arraystretch}{1.}
\renewcommand{\tabcolsep}{5.pt}
\begin{tabular}{c|c|cc|ccc}
\Xhline{2.\arrayrulewidth}
\hline 
\multirow{2}{*}{\textbf{Dataset}} & \multirow{2}{*}{\textbf{Method}}  & \multicolumn{2}{c|}{\textbf{Cls Acc}} & \multicolumn{3}{c}{\textbf{Loc Acc}} \\

& & \textbf{Top-1 } & \textbf{Top-5} &\textbf{Top-1 } & \textbf{Top-5} & \textbf{GT-known} \\ 
\Xhline{2.\arrayrulewidth}
\hline 
\multirow{2}{*}{ Standford Dog~\cite{khosla2011novel}} & TS-CAM & 81.24 & 97.25 & 65.14 & 77.19 & 78.67   \\
 & SAT & \textbf{86.03} ({\color{red} +4.79}) & \textbf{98.61} ({\color{red} +1.36}) &  \textbf{82.97} ({\color{red} +17.83}) & \textbf{94.92} ({\color{red} +17.73}) & \textbf{96.14} ({\color{red} +17.47}) \\
\hline 
\multirow{2}{*}{FGVC-Aircraft~\cite{maji2013fine}} & TS-CAM & 81.28 & 95.41 & 79.69 & 93.22 & 96.73   \\
 & SAT & \textbf{82.66} ({\color{red} +1.38}) & \textbf{95.89} ({\color{red} +0.48}) &  \textbf{82.18} ({\color{red} +2.49}) & \textbf{95.23}({\color{red} +2.01}) & \textbf{98.80} ({\color{red} +2.07}) \\
\hline 
\multirow{2}{*}{Standford Cars~\cite{krause20133d}} & TS-CAM & 83.16 & 96.48 & 79.74 & 92.43 & 95.62  \\
 & SAT & \textbf{85.92} ({\color{red} +2.76}) & \textbf{97.55} ({\color{red} +1.07}) &  \textbf{85.79} ({\color{red} +5.95}) & \textbf{97.35} ({\color{red} +4.98}) & \textbf{99.76} ({\color{red} +4.14}) \\
\hline
\Xhline{2.\arrayrulewidth}
\end{tabular}
\end{center}
\vspace{-2mm}
\caption{\textbf{Fine-grained.} Comparison with TS-CAM method on three fine-grained datasets.}
\label{table:fine_grained}
\end{table*}

\begin{figure*}[t]
\footnotesize
\centering
	\begin{overpic}[width=0.98\linewidth]{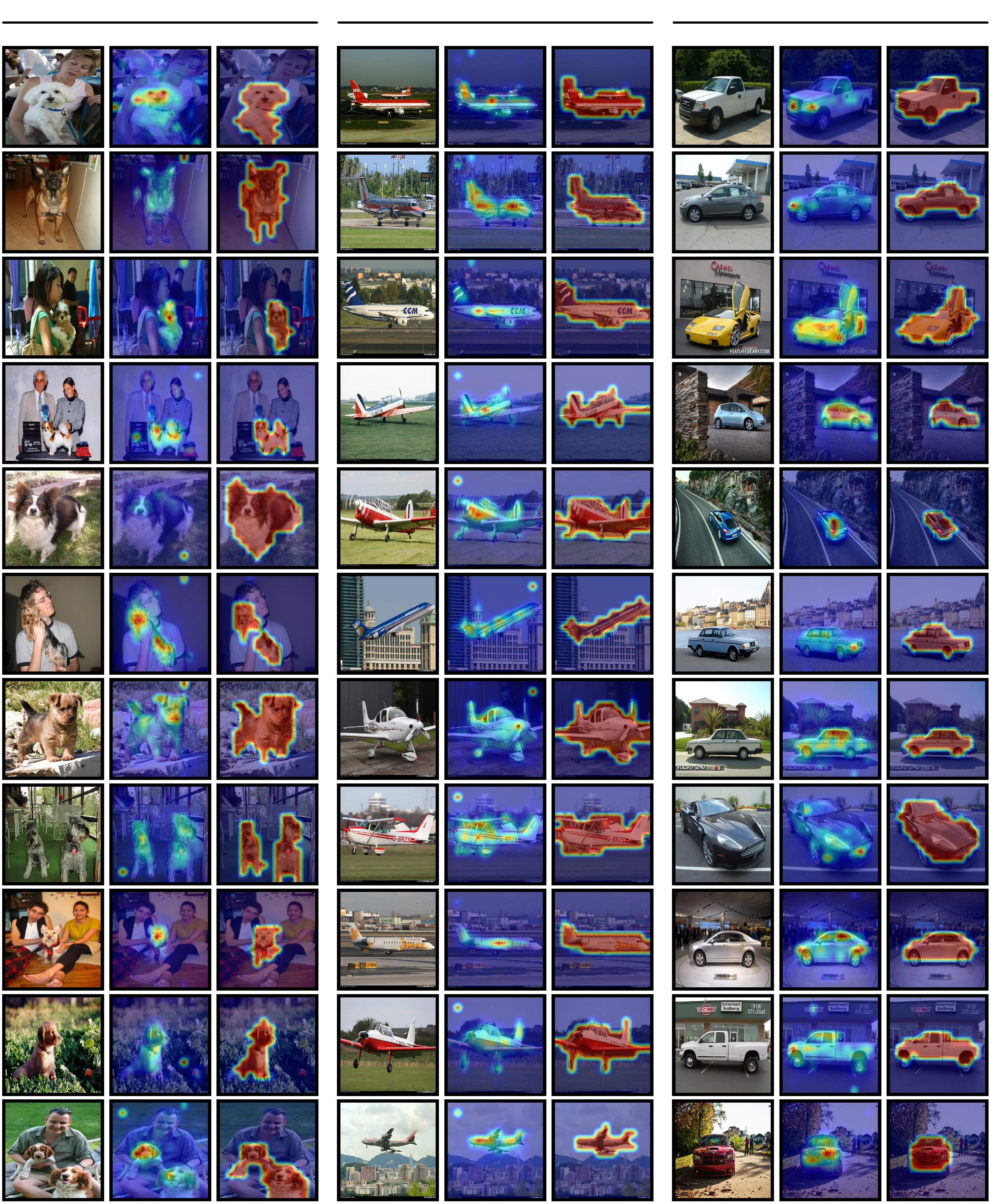}
	    \put(8.3, 98.7){{\textbf{Standford Dogs~\cite{khosla2011novel}}}}
	    \put(3., 96.8){{\textbf{Image}}}
	    \put(10.5, 96.8){{\textbf{TS-CAM}}}
	    \put(20.9, 96.8){{\textbf{Ours}}}
	    
	    \put(36.5, 98.7){{\textbf{FGVC-Aircraft~\cite{maji2013fine}}}}
	    \put(30.5, 96.8){{\textbf{Image}}}
	    \put(38.5, 96.8){{\textbf{TS-CAM}}}
	    \put(48.4, 96.8){{\textbf{Ours}}}
	    
	    \put(64, 98.7){{\textbf{Standford Cars~\cite{krause20133d}}}}
	    \put(58.5, 96.8){{\textbf{Image}}}
	    \put(66.5, 96.8){{\textbf{TS-CAM}}}
	    \put(76.4, 96.8){{\textbf{Ours}}}
	    
\end{overpic}
\caption{Visualization comparison with the baseline TS-CAM~\cite{gao2021ts} method on Standford Dog~\cite{khosla2011novel}, FGVC-Aircraft~\cite{maji2013fine}, and Standford Cars~\cite{krause20133d}.}
\label{fine_visual}
\end{figure*}

\myPara{Fine-grained.}
To further validate the effectiveness of SAT, we compare the accuracy of SAT with TS-CAM~\cite{gao2021ts} on three fine-grained datasets, including Standford Dogs~\cite{khosla2011novel}, FGVC-Aircraft~\cite{maji2013fine}, and Standford Cars~\cite{krause20133d}, as shown in Table \ref{table:fine_grained}. On Standford Dogs, we achieve significant gains of \textbf{17.83\%} and \textbf{17.47\%} on Top-1 Loc and GT-known Loc compared to TS-CAM. Besides, we obtain \textbf{98.80\%} and \textbf{99.76\%} GT-known Loc on FGVC-Aircraft and Standford Cars, exceeding TS-CAM by \textbf{2.07\%} and \textbf{4.14\%}, respectively. Fig.~\ref{fine_visual} illustrates several visual comparisons between TS-CAM and our proposed method on three fine-grained datasets. Compared to TS-CAM, the localization results generated by the proposed method have better visualization and more complete coverage of the object.

\myPara{Comparison with convnet-based methods.} We replace the classifiers of PSOL~\cite{zhang2020rethinking} and SPOL~\cite{wei2021shallow} with Deit-S~\cite{touvron2021training} backbone and report the reproduced localization results in the Table~\ref{table:reproduct}. Compared to the above methods, SAT still achieves the best localization results on both benchmarks.

\begin{table*}[t]
\begin{center}
\renewcommand{\arraystretch}{1}
\renewcommand{\tabcolsep}{8pt}
\centering
\begin{tabular}{c|c|c|c|c|c|c}
\Xhline{2.\arrayrulewidth}
\hline 
\multirow{2}{*}{\textbf{Methods}}&\multicolumn{3}{c|}{\textbf{CUB-200  Loc Acc.}}&\multicolumn{3}{c}{\textbf{ImageNet  Loc Acc.}}\\
\cline{2-7}
    &  \textbf{Top-1} &  \textbf{Top-5} & \textbf{GT-k.} &  \textbf{Top-1} &  \textbf{Top-5} & \textbf{GT-k. }\\
\Xhline{1.\arrayrulewidth}
\hline 
PSOL* &  72.45* & 87.48*  & 90.00  & 54.71* &  63.54* &  65.44 \\
SPOL* &  80.73* &  93.76* &  96.46 & 59.89* & 67.68*  & 69.02\\
\hline
SAT& \textbf{80.96} & \textbf{94.13} & \textbf{98.45} & \textbf{60.15} & \textbf{70.52} & \textbf{73.13}\\
\hline
\Xhline{2.\arrayrulewidth}
\end{tabular}
\end{center}
\vspace{-2mm}
\caption{\textbf{ Reproducing the convnet-based methods} on the Deit-S. * indicates the reproduced results.}
\label{table:reproduct}
\end{table*}

\myPara{Main results.}
In Table~\ref{table:comparsion2}, we show the more complete comparison results with other SOTA methods on CUB-200~\cite{wah2011caltech} and ImageNet~\cite{russakovsky2015imagenet}. It can be seen that the proposed SAT achieves the best performance on both datasets in terms of Top-1/Top-5/GT-known Loc three localization metrics. 


\begin{table*}[t]
\small
\centering
\begin{minipage}[t]{0.98\textwidth}
    \renewcommand{\arraystretch}{1.}
    \renewcommand{\tabcolsep}{3.pt}
    \begin{tabular}{c|cccccc}
    \Xhline{2.\arrayrulewidth}
    \hline 
    
      \textbf{Total Errors}&
      \textbf{Object Occlusion} & \textbf{Localization More} & \textbf{Water Reflection} & \textbf{Localization Part} &
      \textbf{Multiple Instances} & \textbf{Label Error}\\
    \Xhline{2.\arrayrulewidth}
    \hline 
    90 &  36 & 28
     & 18 & 5 & 2 & 1 \\
    \hline
    \Xhline{2.\arrayrulewidth}
    \end{tabular}
\end{minipage}
\centering
    \begin{minipage}[t]{0.98\textwidth}
    \centering
    \begin{overpic}[width=0.99\linewidth]{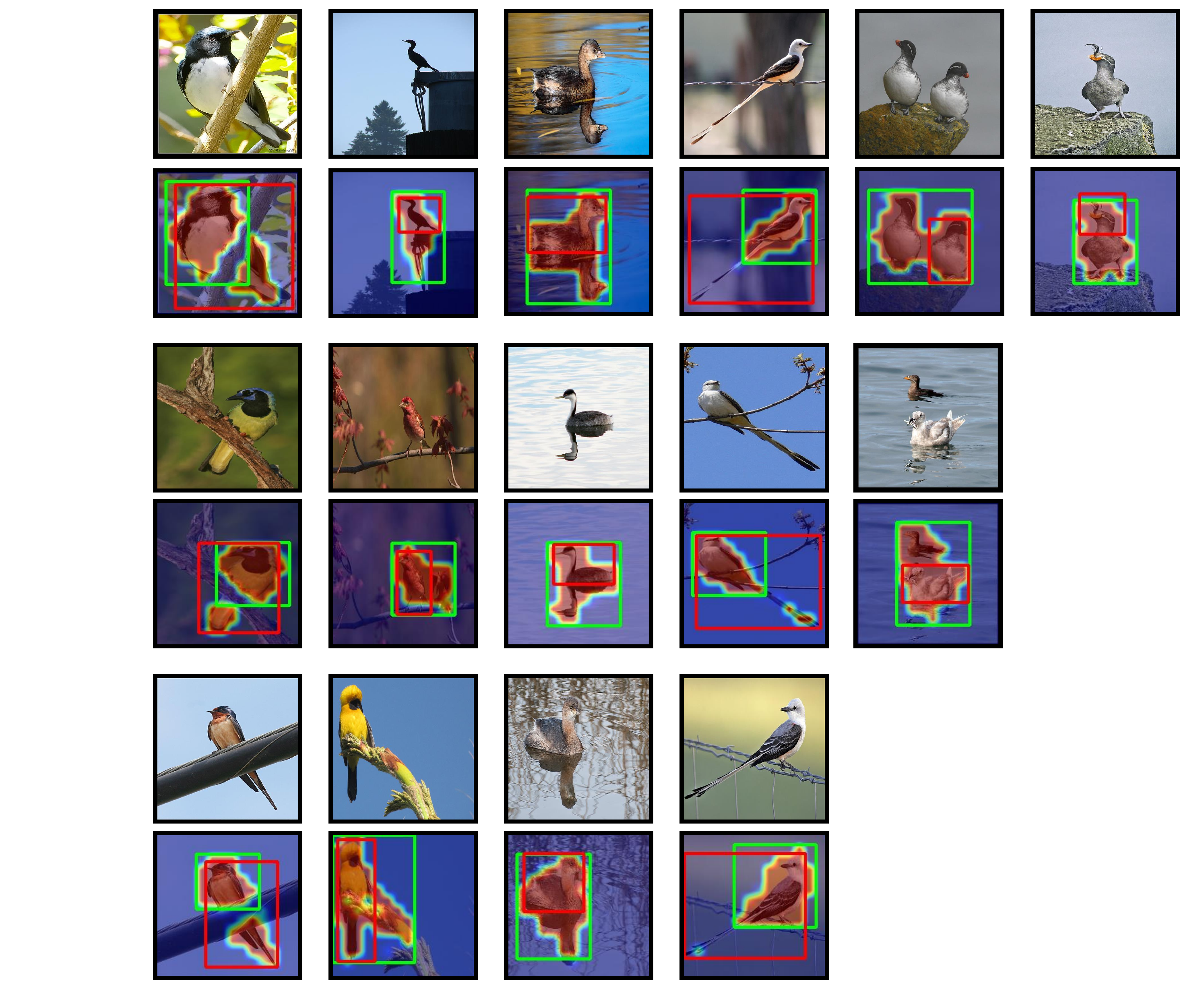}
	    \put(10.5, 4.){\rotatebox{90}{\textbf{Predict}}}
	    \put(10.5, 17.5){\rotatebox{90}{\textbf{Image}}}
	    
	    \put(10.5, 31.5){\rotatebox{90}{\textbf{Predict}}}
	    \put(10.5, 45.){\rotatebox{90}{\textbf{Image}}}
	    
	    \put(10.5, 59){\rotatebox{90}{\textbf{Predict}}}
	    \put(10.5, 72.5){\rotatebox{90}{\textbf{Image}}}
	   
    \end{overpic}
    \caption{\textbf{Localization error analysis} on CUB-200. }
    \label{table:error}
    \end{minipage}
\end{table*}

\begin{table*}[t]
\renewcommand{\arraystretch}{1.}
\renewcommand{\tabcolsep}{7.5pt}
\begin{center}
\small
\begin{tabular}{l|c|c|c|c|c|c|c|c}
\Xhline{2.0\arrayrulewidth}
\hline
\multirow{2}{*}{\textbf{Methods}} & \multirow{2}{*}{\textbf{Venue}} & \multirow{2}{*}{\textbf{Backbone}} &  \multicolumn{3}{c|}{\textbf{CUB-200}~\textbf{\cite{wah2011caltech} Loc Acc.}} & \multicolumn{3}{c}{\textbf{ImageNet}~\textbf{\cite{russakovsky2015imagenet} Loc Acc.}} \\
\cline{4-9}
      & &  & \textbf{Top-1} & \textbf{Top-5} & \textbf{GT-known}  & \textbf{Top-1} & \textbf{Top-5} & \textbf{GT-known} \\
\hline
\Xhline{2.0\arrayrulewidth}
CAM~\cite{zhou2016learning} & CVPR$16$ &VGG16
& 41.06 & 50.66 & 55.10 & 42.80 & 54.86 & 59.00   \\
ACoL~\cite{zhang2018adversarial} & CVPR$18$&VGG16 & 45.92 & 56.51 & 62.96 & 45.83 & 59.43 & 62.96 \\
ADL~\cite{choe2019attention} & CVPR$19$&VGG16
& 52.36 & $-$ & 75.41 & 44.92 & $-$ & $-$\\
DANet~\cite{xue2019danet} & ICCV$19$ &VGG16
&52.52&61.96&67.70 &$-$&$-$&$-$\\
I2C~\cite{zhang2020inter} & ECCV$20$ &VGG16
&55.99&68.34&$-$ & 47.41 & 58.51&63.90\\
MEIL~\cite{mai2020erasing} & CVPR20 &VGG16
& 57.46 & $-$ & 73.84 & 46.81 & $-$ & $-$\\
SLT~\cite{guo2021strengthen} & CVPR$21$ &VGG16
& 67.80 & $-$ & 87.60 & 51.20 & 62.40 & 67.20 \\
ORNet~\cite{xie2021online} & ICCV$21$ &VGG16
& 67.73 & 80.77 & 86.20 & 52.05 & 63.94 & 68.27 \\
BAS~\cite{wu2021background} & CVPR$22$ &VGG16
& 71.33 & 85.33 & 91.07 & 52.96 & 65.41  & 69.64 \\
Kim et al.~\cite{kim2022bridging} & CVPR$22$  &VGG16
& 70.83 & 88.07 & 93.17 & 49.94 & 63.25 & 68.92\\
CREAM~\cite{xu2022cream}  & CVPR$22$ & VGG16
& 70.44 &85.67 & 90.98 & 52.37 & 64.20 & 68.32 \\
\hline
CAM~\cite{zhou2016learning} & CVPR$16$ & InceptionV3
&41.06&50.66&55.10 & 46.29&58.19&62.68\\
SPG~\cite{zhang2018self} & ECCV$18$ &InceptionV3
& 46.64 & 57.72 & $-$ & 48.60 & 60.00 & 64.69 \\
DANet~\cite{xue2019danet} & ICCV$19$ &InceptionV3
& 49.45 & 60.46 & 67.03 & 47.53 & 58.28 & $-$\\
I2C~\cite{zhang2020inter} & ECCV$20$ &InceptionV3
& 55.99 & 68.34 & 72.60 & 53.11 & 64.13 & 68.50 \\
GCNet~\cite{lu2020geometry} & ECCV$20$ &InceptionV3
& 58.58 & 71.00 & 75.30 & 49.06 & 58.09 & $-$\\
SPA~\cite{pan2021unveiling} & CVPR$21$ &InceptionV3
& 53.59 & 66.50 & 72.14 & 52.73 & 64.27 & 68.33 \\
FAM~\cite{meng2021foreground} & ICCV$21$ &InceptionV3
& 70.67 & $-$ & 87.25 & 55.24 & $-$ & 68.62 \\
CREAM~\cite{xu2022cream} & CVPR$22$ &InceptionV3 & 71.76 & 86.37 & 90.43 & 56.07 & 66.19 & 69.03 \\
BAS~\cite{wu2021background} & CVPR$22$ &InceptionV3
& 73.29 & 86.31 & 92.24 & 58.51 &\underline{69.00} & 71.93 \\
BagCAMs~\cite{zhu2022bagging} & ECCV$22$ & InceptionV3  & 60.07  & $-$ & 89.78 & 53.87  & $-$  & 71.02 \\
\hline
CAM~\cite{zhou2016learning} & CVPR$16$ &ResNet50
&46.71&54.44&57.35 &38.99&49.47&51.86\\
ADL~\cite{choe2019attention} & CVPR$19$ &ResNet50
&62.29&$-$&$-$ &48.53&$-$&$-$\\
I2C~\cite{zhang2020inter} & ECCV$20$ &ResNet50
&$-$&$-$&$-$ & 51.83&64.60&68.50\\
PSOL~\cite{zhang2020rethinking} & CVPR$20$ &ResNet50
& 70.68 & 86.64 & 90.00 & 53.98 & 63.08 & 65.44 \\
FAM~\cite{meng2021foreground} & ICCV$21$ &ResNet50
& 73.74 & $-$ & 85.73 & 54.46 & $-$ & 64.56 \\
SPOL~\cite{wei2021shallow} & CVPR$21$&ResNet50
& 80.12 & 93.44 & 96.46 & 59.14 & 67.15 & 69.02 \\
DA-WSOL~\cite{zhu2022weakly} & CVPR$22$ &ResNet50
& 66.65 & $-$ & 81.83 & 55.84 & $-$ & 70.27 \\
BAS~\cite{wu2021background} & CVPR$22$ &ResNet50
& 77.25 & 90.08 & 95.13 & 57.18 & 68.44 & 71.77 \\
Kim et al.~\cite{kim2022bridging} & CVPR$22$  &ResNet50
& 73.16 & 86.68 & 91.60 & 53.76 & 65.75 & 69.89\\
CREAM~\cite{xu2022cream} & CVPR$22$ & ResNet50
& 76.03 & $-$ & 89.88 & 55.66 & $-$ &  69.31 \\
BagCAMs~\cite{zhu2022bagging} & ECCV$22$ & ResNet50  & 69.67  & $-$ & 94.01 & 44.24  & $-$  & \underline{72.08} \\
ISIC~\cite{wei2022weakly} & ECCV$22$ & ResNet50  & \underline{80.68} & \underline{94.08} & \underline{97.32} & \underline{59.61}  & 67.84  & 70.01 \\
\hline
TS-CAM~\cite{gao2021ts} & ICCV$21$ &Deit-S
& 71.30 & 83.80 & 87.70 & 53.40 & 64.30 & 67.60 \\
LCTR~\cite{chen2021lctr} & AAAI$22$ &Deit-S
& 79.20 & 89.90 & 92.40 & 56.10 & 65.80 & 68.70 \\
SCM~\cite{bai2022weakly} & ECCV$22$ &Deit-S
& 76.40 & 91.60 & 96.60 & 56.10 & 66.40 & 68.80 \\
\hline
\rowcolor{mygray}

SAT ({\color{red} ours}) & This Work & Deit-S & \textbf{80.96} & \textbf{94.13} & \textbf{98.45} & \textbf{60.15} & \textbf{70.52} & \textbf{73.13}\\ 

\hline
\Xhline{2.0\arrayrulewidth}
\end{tabular}
\end{center}
\vspace{-2mm}
\caption{\textbf{Comparison with state-of-the-art methods.} The best results are highlighted in \textbf{bold}, second are \underline{underlined}.}
\label{table:comparsion2}
\end{table*}

\begin{figure*}[t]
\footnotesize
\centering
	\begin{overpic}[width=0.98\linewidth]{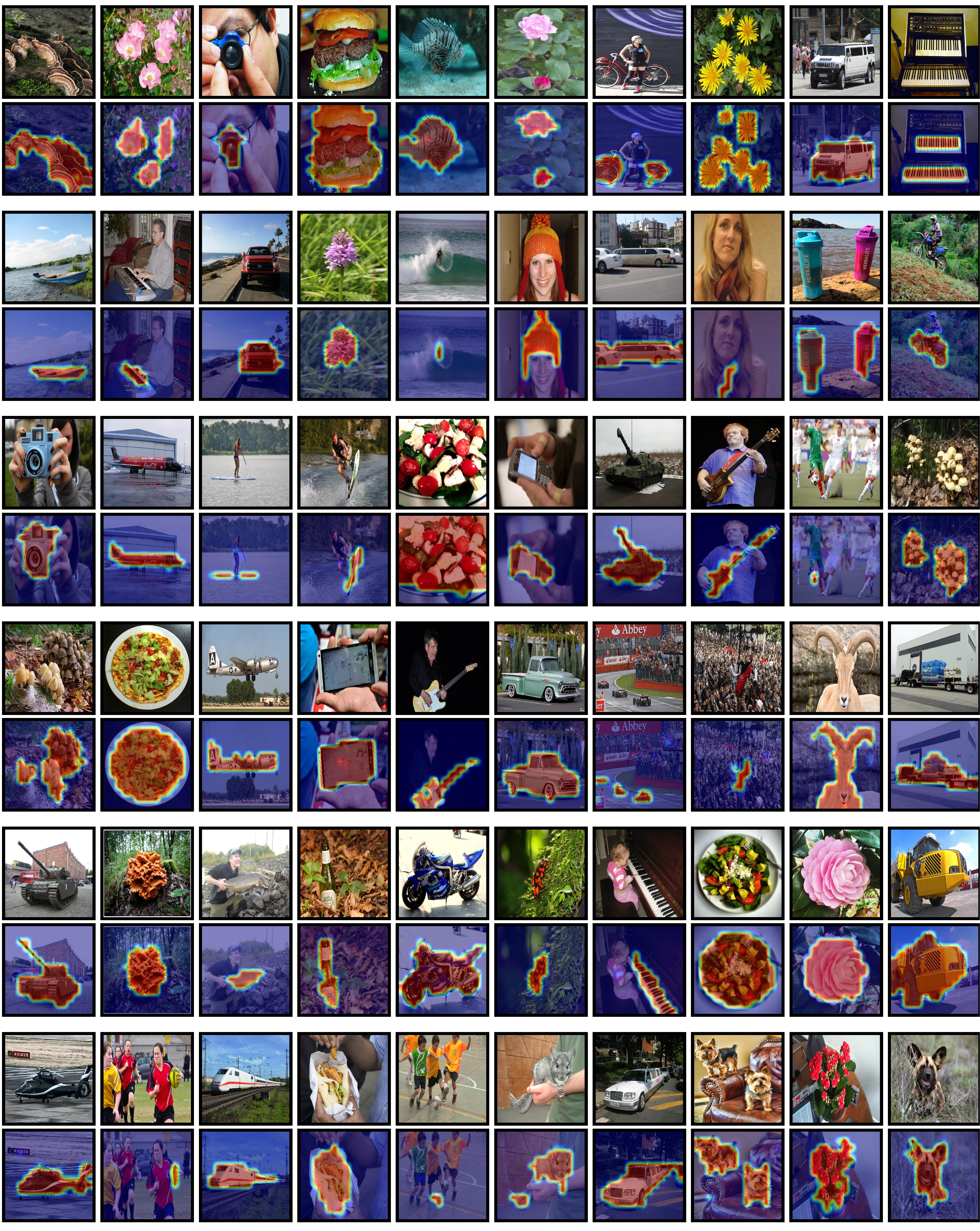}
\end{overpic}
\caption{Visualization of the localization results on OpenImages~\cite{choe2020evaluating}.}
\label{openimages_supp}
\end{figure*}

\begin{figure*}[t]
\footnotesize
\centering
	\begin{overpic}[width=0.98\linewidth]{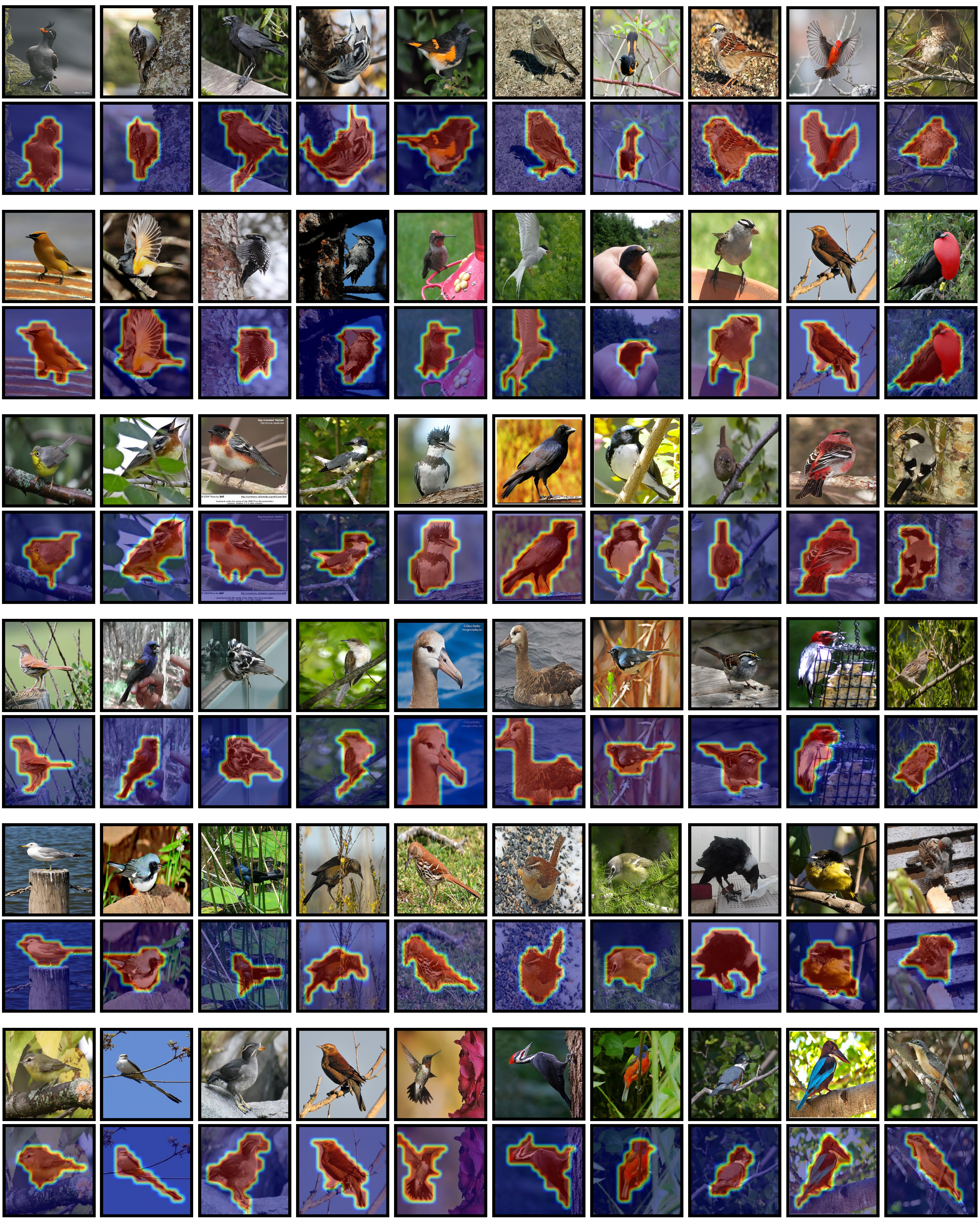}
\end{overpic}
\caption{Visualization of the localization results on CUB-200~\cite{russakovsky2015imagenet}.}
\label{cub_supp}
\end{figure*}

\begin{figure*}[t]
\footnotesize
\centering
	\begin{overpic}[width=0.98\linewidth]{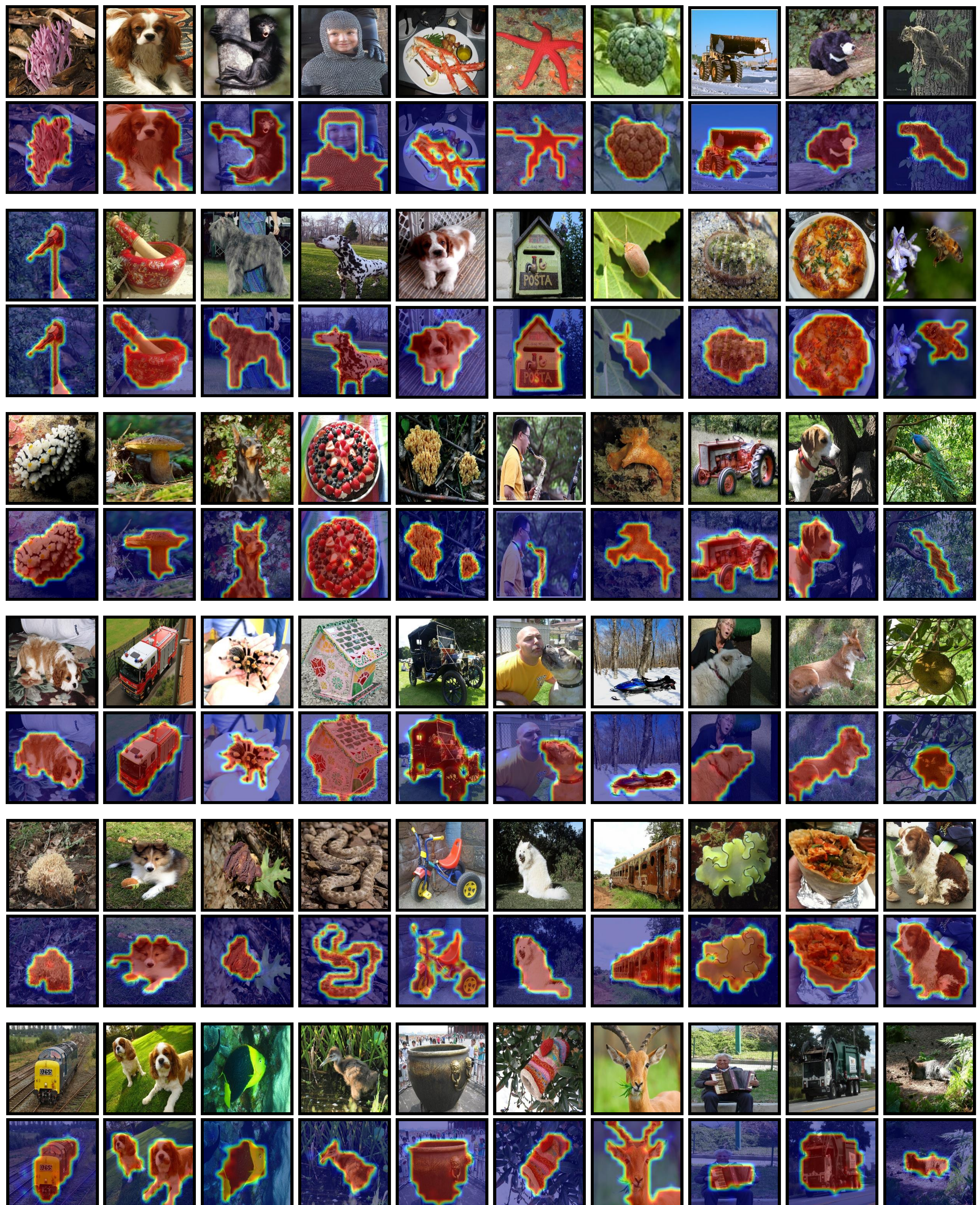}
\end{overpic}
\caption{Visualization of the localization results on ImageNet~\cite{wah2011caltech}.}
\label{imgenet_supp}
\end{figure*}

\myPara{Visual Results.}
More visualizations on OpenImages ~\cite{choe2020evaluating}, CUB-200~\cite{wah2011caltech}, and ImageNet~\cite{russakovsky2015imagenet} datasets are shown in Fig.~\ref{openimages_supp}, Fig.~\ref{cub_supp}, and Fig.~\ref{imgenet_supp}, respectively. It can be noted that SAT demonstrates robust localization ability in various challenging scenarios, including different scaled objects, complex environments, and object occlusions.

\end{document}